%% file: main.tex
\documentclass{article}
\usepackage{ijcai26}

\usepackage{times}
\usepackage{soul}
\usepackage{url}
\usepackage[hidelinks]{hyperref}
\usepackage[utf8]{inputenc}
\usepackage[small]{caption}
\usepackage{graphicx}
\usepackage{amsmath}
\usepackage{amsthm}
\usepackage{amsfonts}
\usepackage{booktabs}
\usepackage{algorithm}
\usepackage{algorithmic}
\usepackage[switch]{lineno}

\usepackage{array}

\usepackage{xcolor}

\title{\textit{UPath}:\\\textit{U}niversal \textit{P}lanner \textit{A}cross \textit{T}opological \textit{H}eterogeneity For Grid-Based Pathfinding}

\author{
Aleksandr Ananikian\thanks{Equal contribution.}
\and
Daniil Drozdov\footnotemark[1]\and
Konstantin Yakovlev\\
\affiliations
Saint-Petersburg University\\
\emails
\{a.ananikian, d.drozdov, k.yakovlev\}@spbu.ru,
}

\begin{document}

\maketitle

\begin{abstract}

The performance of search algorithms for grid-based pathfinding, e.g. A*, critically depends on the heuristic function that is used to focus the search. Recent studies have shown that informed heuristics that take the positions/shapes of the obstacles into account can be approximated with the deep neural networks. 
Unfortunately, the existing learning-based approaches mostly rely on the assumption that training and test grid maps are drawn from the same distribution (e.g., city maps, indoor maps, etc.) and perform poorly on out-of-distribution tasks. This naturally limits their application in practice when often a universal solver is needed that is capable of efficiently handling any problem instance. In this work, we close this gap by designing an \textit{universal heuristic predictor}: a model trained once, but capable of generalizing across a full spectrum of unseen tasks. Our extensive empirical evaluation shows that the suggested approach halves the computational effort of A* by up to a factor of $2.2$, while still providing solutions within $3\%$ of the optimal cost on average altogether on the tasks that are completely different from the ones used for training -- a milestone reached for the first time by a learnable solver.


\end{abstract}

\section{Introduction}

Path planning in static or unknown environments is a fundamental and extensively studied problem in AI and robotics. A common formulation discretizes a continuous workspace into a grid, frequently an 8‑connected one, and reduces navigation to a shortest‑path search task. In such graphs, nodes represent grid cells and edges encode permissible transitions.

\begin{figure}[t]
  \centering
  \includegraphics[width=1\linewidth]{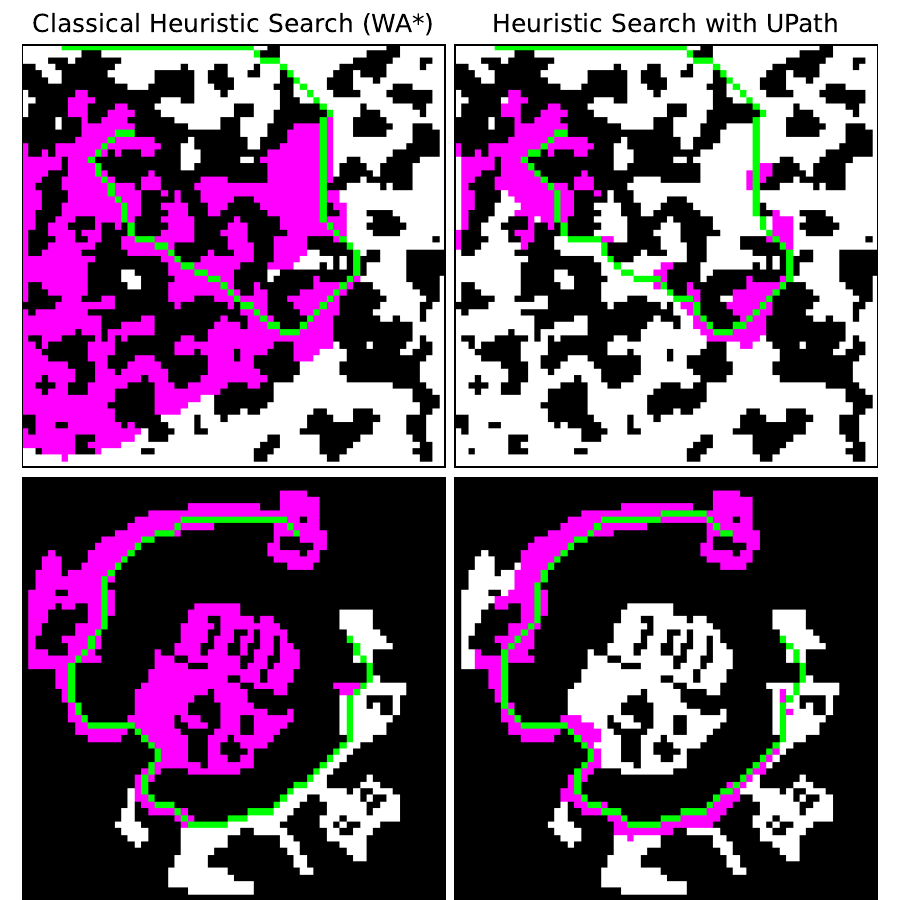}
\caption{The difference between WA* and our approach. Expanded nodes are shown in magenta, while path in green.}
\label{fig:visual-abstract}
\end{figure}

Heuristic search algorithms, such as A* \cite{Hart1968}, remain the algorithmic backbone for grid‑based planning due to their elegance, completeness, and tunable efficiency. However, their performance critically depends on the quality of the heuristic -- an estimate of cost‑to‑go from each node to the goal. While Manhattan or octile distance heuristics are standard, they are instance‑independent and fail to exploit the obstacle layout of a particular environment, often leading to excessive node expansions especially in cluttered maps.

To overcome these limitations, recent research has explored deep learning for constructing instance-aware heuristics. By viewing grid maps as binary images, convolutional neural networks, sometimes augmented with attention or transformers, are trained to produce cost-to-go predictions or guidance maps that influence search algorithms. Notably, methods like Neural A* \cite{pmlr-v139-yonetani21a}
reformulate A* in a differentiable end‑to‑end framework, improving path optimality and search efficiency compared to classical heuristics. Another representative method is TransPath \cite{Kirilenko_Andreychuk_Panov_Yakovlev_2023}, which learns a correction factor -- the ratio between a standard heuristic for 8-connected grid (i.e. the octile distance) and the perfect heuristic that equals the cost of the shortest path to a goal cell on a given grid. This approach has demonstrated up to $4$× reduction in node expansions while producing solutions within $0.3\%$ of optimal cost.

Despite these advances, a fundamental limitation remains: most learning-based heuristics are developed under an implicit \emph{in-distribution} assumption, where training and deployment maps share the same generative structure. This assumption is often violated in practice, where planners must operate across heterogeneous environments and under distribution shift, typically without the opportunity to retrain or re-tune the model for each new domain.

In this work, we pursue \emph{train once, search everywhere} paradigm by proposing a universal heuristic predictor -- a single neural network trained once and then used as a drop-in heuristic module for classical search across a broad spectrum of grid-based planning tasks, including strictly out-of-distribution scenarios. Our solver consists of two components. First, given an input grid, the network predicts a dense correction factor map $cf(\cdot)$, where for each cell $n$, $cf(n)$ is defined as the ratio between a standard geometric heuristic (octile distance) and the perfect heuristic (the true shortest-path cost-to-go on the given map). Second, we run a heuristic search procedure (A*) using the resulting heuristic values. Predicting a correction factor, rather than regressing the absolute cost-to-go, retains a strong geometric prior while allowing the model to account for obstacle-induced detours.

Crucially, our training protocol does not rely on injecting representative out-of-distribution examples. To study this regime explicitly, we train only on maps generated from simple stochastic priors (including purely local randomness, global density variation, and structured obstacle patches), but evaluate on a separate suite of $20{,}000$ planning tasks drawn from ten qualitatively different topology generators spanning both realistic layouts and diverse synthetic patterns. Supervision is obtained by running Dijkstra's algorithm from the goal to compute the perfect cost-to-go, from which correction-factor targets are derived; obstacle and goal cells are masked during training to avoid degenerate supervision.

Empirically, compared to vanilla A*, our method reduces computational effort by up to a factor of 2.2 while producing solutions within approximately 3\% of optimal cost on average. It also outperforms Weighted A* and prior learnable planners, yielding a more favorable efficiency--generalization trade-off. Figure~\ref{fig:visual-abstract} qualitatively illustrates this effect: across maps with different spatial structure, our learned guidance concentrates expansions into decision-relevant regions while still recovering a valid route, in contrast to the broader expansion patterns induced by heuristic inflation.

\section{Related Work}
The research areas most relevant to this work are learning for search -- specifically, learning for pathfinding -- and the evolution of pathfinding datasets used to study navigation and shortest-path planning.

\paragraph{Learning-guided best-first search}
Many planning and reasoning problems can be cast as search on a (possibly implicit) state-transition graph, unifying classical path planning with combinatorial puzzles, logic synthesis, and retrosynthetic planning \cite{NEURIPS2024_bc8b2058}. A large fraction of practical approaches rely on best-first search (BestFS), where an evaluation function $f$ prioritizes which frontier state to expand next. Learning-based methods improve search by predicting effective components of $f$, such as heuristics, achieving strong results in domains like Rubik's Cube \cite{agostinelli2019solving} and Sokoban \cite{orseau2023ltscm}. Recent work further explores training guidance that transfers across multiple environments \cite{ijcai2024p743}.

\paragraph{Machine learning for pathfinding}
Within pathfinding, learning-based methods can be roughly grouped into two main directions. The first (and most closely related to our work) learns a heuristic or search guidance signal that can be used at inference time within a classical planner such as A*. The latter remains unchanged, but its expansion order is shaped by a learned model, enabling plug-and-play integration and preserving the desirable guarantees and modularity of classical search. Representative examples include approaches that predict heuristics for A*-like algorithms SaIL \cite{pmlr-v78-bhardwaj17a}, TransPath \cite{Kirilenko_Andreychuk_Panov_Yakovlev_2023}.

The second direction focuses on differentiable planning, where gradients are propagated through a planning procedure to train neural components end-to-end. Examples include \cite{pmlr-v139-yonetani21a} and \cite{chen2025iastar}. These methods are conceptually appealing because they directly optimize planning behavior, but they are often challenging to deploy in practice due to slow training, instability, and sensitivity to hyperparameters, especially when scaling to large environments or long-horizon searches. 

\paragraph{Path planning datasets and evaluation distributions}
Benchmarking grid-based path planning has traditionally relied on simulated environments. A widely used reference collection is the Moving AI Lab benchmark suite \cite{sturtevant2012benchmarks}, which synthetic maps (e.g., mazes, rooms, and random obstacles), maps derived from commercial games such as \emph{Baldur's Gate}, and large-scale city/street layouts. Another common source is the Motion Planning (MP) dataset \cite{pmlr-v78-bhardwaj17a}, which provides eight families of grid environments with distinctive obstacle patterns. Building on MP, the Tiled Motion Planning (TMP/TiledMP) dataset \cite{pmlr-v139-yonetani21a} composes multiple MP maps into larger layouts to increase structural diversity. Despite their utility, these benchmarks still only partially reflect the spatial statistics of real indoor environments. HouseExpo \cite{Li2019HouseExpo}, based on the manually created SUNCG dataset \cite{song2016ssc}, was introduced to narrow this gap by providing a large-scale collection of 2D indoor layouts. In our work, we sample a subset from each of the above datasets and rescale all maps to a fixed grid resolution, as well as we add a range of other procedurally generated layouts that cover different grid topologies.

\section{Background}
\subsection{Pathfinding problem}
Consider a graph $G$, specifically, a grid made up of the blocked and free cells. From any free cell, an agent can move to any of its eight neighboring cells, four in the cardinal directions, and four diagonally, provided that the destination cell is also free. Movements in the cardinal directions each carry a cost of $1$, while diagonal moves cost $\sqrt{2}$. This framework is known as an 8-connected grid with non-uniform movement costs.

A path between two distinguished free cells $start$ and $goal$ is a sequence of the adjacent cells connecting them: $\pi(start, goal) = (c_0=start, c1, c2, \ldots, c_n = goal)$. This path is valid if and only if every cell $c_i$ in the sequence is free. The cost of a valid path is

\[
\mbox{cost}(\pi) = \sum_{i=0}^{n-1} \mbox{cost}(c_i, c_{i+1}).
\]

Let $\Pi$ denote the set of all valid paths from $start$ to $goal$. The optimal (minimum-cost) path $\pi^*$ is the one that satisfies $\forall \pi \in \Pi: \: \mbox{cost}(\pi^*) \leq \mbox{cost}(\pi)$.

Now, the pathfinding problem (on an 8-connected grid) can be formally stated as a triplet: $\textbf{P} = (G, start, goal)$. To solve this problem means to construct a corresponding valid path. An optimal solution is a shortest path $\pi^*$. 

In this paper, we focus on finding valid paths efficiently, rather than constructing the optimal ones.

\subsection{A* search}

A* is a heuristic search algorithm which is widely used to solve the pathfinding problems described above and is renowned for its simplicity and strong theoretical guarantees. During execution, A* incrementally grows a search tree whose nodes represent grid cells augmented with search metadata. Typically, each node stores the $g$-value, which is the cost of path from the start node to this node, the $h$-value, a heuristic estimate of the cost from this node to the goal, the $f$-value, which is the sum of $g$- and $h$-values.

At each iteration, A* selects a node with the smallest $f$-value from the search frontier, called OPEN, to be expanded. Expanding a node is comprised of generating all its valid successors, those reachable by a legal move on the grid, calculating the $g$-value of each successor by adding the appropriate transition cost to the parent’s $g$-value, and then updating the search tree appropriately. Specifically, if a successor is a newly generated node then it is immediately added to the tree as the leaf; if a tree already contains a similar node but with the better $g$-value, the successor is discarded; if a tree contains a similar node but its $g$-values is worse (i.e. a better path to this node is found) then the latter is updated. Each newly added or updated search node is added to OPEN. The expanded node is removed from the search frontier and is marked as CLOSED. The algorithm terminates when the goal node is removed from OPEN, at which point the sought path can be reconstructed by following the back-pointers in the search-tree.

The crucial component of A* algorithm is a heuristic function that can be considered as an input to the algorithm (along with the specific pathfinding problem instance to solve). Specifically, how well this function estimates the path costs influences vastly on the number of expansions/iterations and on the quality of the output solution, i.e. the cost of the resultant path.

\paragraph{Heuristics}

A perfect heuristic $h^*$ always gives the exact cost-to-go: $h^*(n) = \mbox{cost}(\pi^*(n, goal))$. An admissible heuristic never overestimates this cost: $\forall n: \: h(n) \leq h^*(n)$. A consistent (or monotone) heuristic satisfies $\forall n, n': \: h(n) \leq h(n') + \mbox{cost}(\pi^*(n, n'))$, which implies admissibility and guarantees that each node is expanded at most once. All of these heuristics result in finding optimal solutions.

For 8-connected grids, a common admissible and consistent heuristic is the Octile heuristic:
\[
h_{\mathrm{oct}}(n)=\sqrt{2}\min(\Delta x,\Delta y)+|\Delta x-\Delta y|,
\]
where $\Delta x=|x_n-x_g|$ and $\Delta y=|y_n-y_g|$, with $g$ denoting the goal node. Unfortunately, this heuristic is overly general and do not take the specifics of each pathfinidng problem instance into account, i.e. the locations and shapes of the obstacles (blocked grid cells). In practice this often results in guiding the search into the regions near obstacles, expanding many nodes before settling on the shortest path. Thus, one way to improve the search efficiency is to develop and utilize dedicated preprocessing techniques that produce more accurate cost‑to‑go estimates (that is, closer to $h^*$), so that nodes off the optimal route acquire higher heuristic values and are less likely to be expanded. This is the way we follow in this study.

\subsection{Weighted A*}

A common technique for trading off solution optimality against runtime in grid-based pathfinding is to use a weighted heuristic. Instead of ranking the nodes in OPEN by $f(n) = g(n) + h(n)$, Weighted A* (WA*) orders them by $f_w(n) = g(n) + w \cdot h(n)$, where the weight, $w \geq 1$, is specified by the user. This simple adjustment is likely to notably reduce the number of search iterations in practice, while allowing finding the solutions whose cost is at most $w$ times the optimal ones. In practice, WA* is widely used when faster search is more important than perfect optimality, especially since many real‑world applications can tolerate a small cost increase in exchange for significantly fewer node expansions.

\section{Problem Statement}

Consider a fixed heuristic search algorithm \texttt{Alg} that takes as input a grid-based path finding problem $\mathbf{P}=(G, start, goal)$ and a heuristic function $h_{base}$, used to guide the search, to produce a valid path $\pi(start,goal)$. Consider now a distribution of all possible path finding problems, where the input grid has a fixed size, $\mathcal{P}$. In this work we focus on $64 \times 64$ grids aligning with much of the previous work on learing-based pathfinding. Let $h_{\theta}$ be a heuristic function represented as a neural network with parameters $\theta$. 

Let $\mathbf{P} \sim \mathcal{P}$ be an arbitrary pathfinding problem and denote by $\mathrm{exp}(\texttt{Alg},h,\textbf{P})$ the number of nodes expanded by \texttt{Alg} when guided by $h$. Similarly, let $\mathrm{cost}(\texttt{Alg},h,\mathbf{P})$ denote the cost of the corresponding solution. Utilizing $\mathrm{exp}$ and $\mathrm{cost}$ the following normalized criteria can be contstucted:

\[
f_1(\theta,\textbf{P})
:=
\mathrm{CostRatio}(h_{\theta},\textbf{P})
=
\frac{\mathrm{cost}(\texttt{Alg},h_{\theta},\textbf{P})}{\mathrm{cost}(\texttt{Alg},h_{base}, \textbf{P})},
\]
\[
f_2(\theta,\textbf{P})
:=
\mathrm{ExpRatio}(h_{\theta},\textbf{P})
=
\frac{\mathrm{exp}(\texttt{Alg},h_{\theta},\textbf{P})}{\mathrm{exp}(\texttt{Alg},h_{base}, \textbf{P})}.
\]

Informally $f_1$ and $f_2$ tell how \texttt{Alg} equipped with $h_{\theta}$ is better (or worse) compared to the baseline (i.e., \texttt{Alg} with $h_{base}$) in terms of solution cost and expansions respectively. The weighted combination of $f_1$ and $f_2$ is also of particluar interest:

\[
J_{\theta,\textbf{P}}(\lambda) := (1-\lambda)\,f_1(\theta,\textbf{P}) + \lambda\,f_2(\theta,\textbf{P}),
\qquad \lambda \in [0,1],
\]

This objective captures the trade-off between solution cost and computational efficiency (measured as the number of expansions). Our goal and is to minimize its expectation over tasks:

\[
\tilde{\theta}
=
\arg\min_{\theta}\;
\mathbb{E}_{\textbf{P} \sim \mathcal{P}}\big[J_{\theta;\textbf{P}}(\lambda)\big].
\]

Informally, we wish to learn a neural network representing a heuristic function so that the latter helps the search algorithm in finding the solutions of acceptable cost faster than the baseline algorithm. In our work we use \texttt{A$^*$} with $h_{oct}$ as the baseline and we approximate the expectation with the empirical average over a finite dataset $\mathcal{D}$ of test tasks that we carefully design.

\section{Method}
\label{sec:method}
Our solver, named UPath, is naturally a heuristic predictor for grid-based pathfinding under strong topological shift, hence \emph{U} -- universal. It relies on four key components:  a correction-factor formulation of the heuristic function; an encoder--transformer--decoder network with long skip connections and specific loss for accurate dense prediction; a training dataset generated solely from simple procedural priors that prevents the network from overfittiting. Finally, an important contribution is a novel dataset of test instances that is tailored to empirically evaluate solvers under a full spectrum of grid topologies -- an universal evaluation suite (UPF).

\subsection{Heuristic function and search procedure}
\label{sec:heuristic}

Correction factor is combination of the baseline heuristic and the ideal heuristic, as suggested in~\cite{Kirilenko_Andreychuk_Panov_Yakovlev_2023}.
\begin{equation}
  \mathrm{cf}^*(n) = \frac{h_{\text{oct}}(n)}{h^*(n)} .
  \label{eq:cf_target}
\end{equation}

It is this heuristic that we wish our network to predict. To compute it for each problem (at the training phase) we run Dijkstra from the goal to obtain $h^*(n)$ for all free cells and then compute $\mathrm{cf}^*(n)$ via \eqref{eq:cf_target} wherever it is well-defined. Specifically, for reachable non-goal cells, $0 < \mathrm{cf}^*(n) \le 1$ (as $h_{\text{oct}}$ is admissible).  For unreachable cells, $\mathrm{cf}^*(n)=0$ as $h^*(n)=\infty$.  For obstacle cells and the goal cell (where $h_{\text{oct}}=h^*=0$), the target in \eqref{eq:cf_target} is undefined. We therefore exclude such cells from the training loss via masking (described later in Section~\ref{sec:loss}), and set their stored target to a dummy constant (i.e., $1$) purely for implementation convenience.



At test time the network outputs a dense map $\widehat{\mathrm{cf}}(n)\in(0,1]$ that we convert into a regular heuristic  suitable for \texttt{A$^*$}:
\begin{equation}
  \widehat{h}(n) = \frac{h_{\text{oct}}(n)}{\max(\widehat{\mathrm{cf}}(n),\varepsilon)} ,
  \label{eq:h_from_cf}
\end{equation}
with a small $\varepsilon=10^{-9}$ for numerical stability.

Overall, the planner remains standard, only the heuristic is learned.

\begin{figure}[t]
    \centering
    \includegraphics[width=0.45\columnwidth]{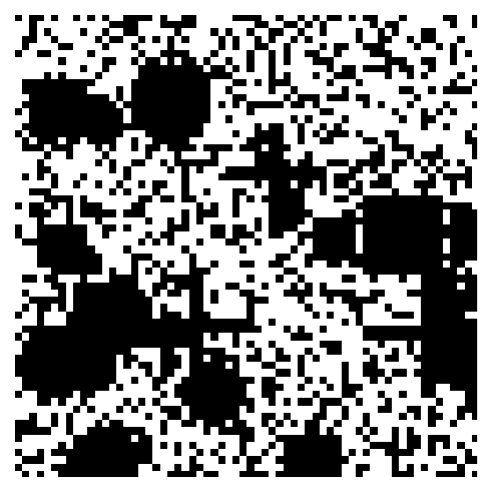}
    \includegraphics[width=0.45\columnwidth]{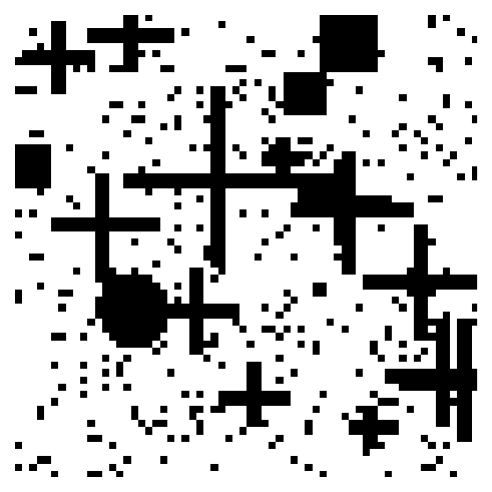}\\
    \includegraphics[width=0.45\columnwidth]{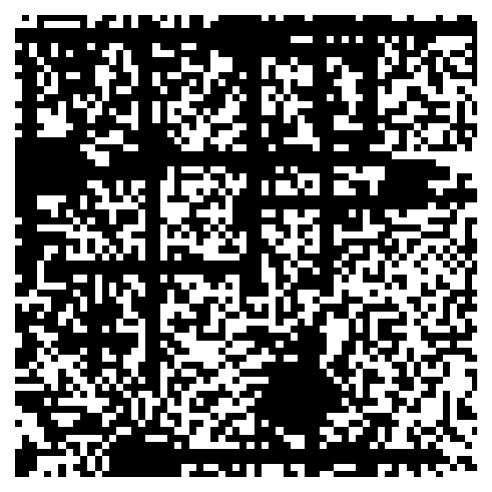}
    \includegraphics[width=0.45\columnwidth]{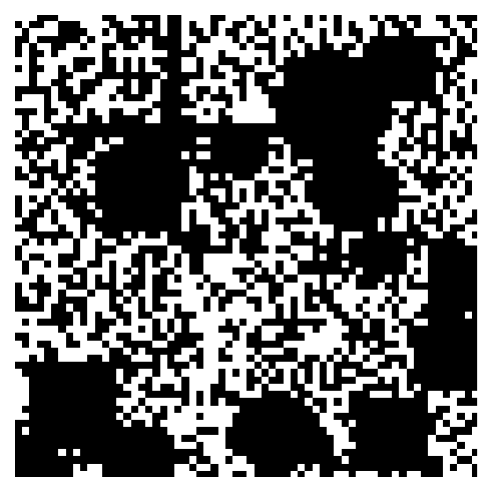}
    \caption{Beta-Figures, 64x64.}
    \label{fig:two-plus-one}
\end{figure}

\begin{figure*}[t]
    \centering
    \includegraphics[width=\textwidth]{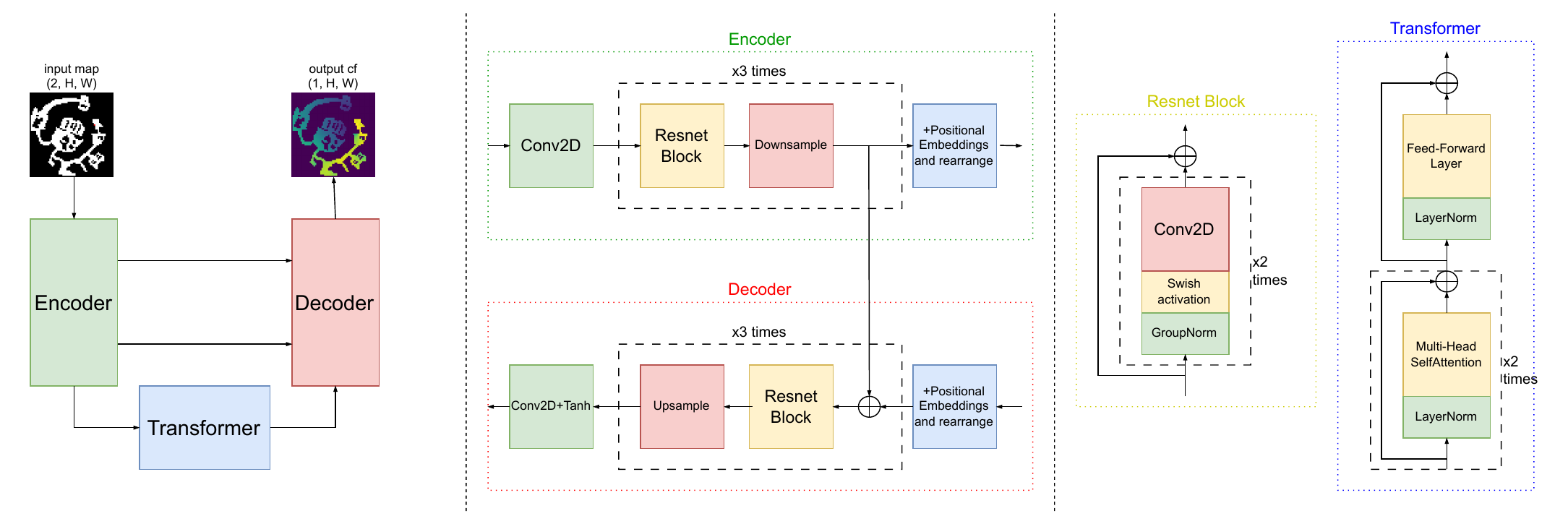}
    \caption{Neural network architecture. }
    \label{fig:arch}
\end{figure*}

\subsection{Training Datasets}
\label{sec:trainsets}

In contrast to prior work in learning-based pathfinding that is centered around utilizing for training the tasks sampled from the same (limited) distribution as for testing, we aim to learn a universal predictor that being trained once is capable to accurately predict the heuristic and, thus, accelerate search for \emph{an arbitrary} pathfinding problem. In other words, we strive for a highly generaliziable learnable heuristic function. To this end we suggest to avoid any specific grid topologies for training but rather rely on simplistic geometric priors -- random noise and basic shapes. As confirmed by our experiments this, indeed, helps us achieve our goal. Overall, we use the following grids for training.

\paragraph{Uniform.}
Each cell is independently traversable with probability $p=0.5$, producing high-frequency random obstacle patterns.

\paragraph{Beta.}
We introduce per-map variability in obstacle density: sample $\theta\sim \mathrm{Beta}(2,2)$ and mark each cell as \emph{blocked} independently with probability $\theta$.
Conditioned on $\theta$, cells are independent, but the marginal obstacle count follows a beta-binomial distribution, yielding substantially higher map-to-map variance (i.e., more very sparse and very dense maps) than the fixed-density Uniform generator.
For example, on a $64\times 64$ grid the probability of generating an entirely empty map is
$\mathbb{E}[(1-\theta)^{4096}] = \frac{6}{(4098)(4099)} \approx 3.6\times 10^{-7}$,
whereas under Uniform($p{=}0.5$) it is $2^{-4096}\approx 10^{-1233}$.

\paragraph{Beta-Figures.}
We inject explicit spatial structure using geometric obstacle primitives -- see Fig.~\ref{fig:two-plus-one}.
Let $M_{\text{figures}}$ be a binary mask formed by randomly placed and sized shapes (e.g., circles, squares, crosses), each occupying multiple cells (at least 10), and let $M_{\beta}$ be a stochastic background mask sampled as in the Beta dataset.
We define the final obstacle map by conjunction:
\begin{equation}
    M_{\text{final}} \;=\; M_{\text{figures}} \wedge M_{\beta}.
\end{equation}
This produces coherent obstacle regions while retaining stochastic variability in density and fragmentation.

\subsection{Neural heuristic predictor}
\label{sec:model}

The network takes an input tensor of shape $(2,H,W)$: an obstacle indicator channel and a goal-indicator channel, and outputs a dense correction-factor map $\widehat{\mathrm{cf}}\in\mathbb{R}^{H\times W\times 1}$.


We follow the encoder--transformer--decoder backbone popularized by state-of-the-art TransPath model~\cite{Kirilenko_Andreychuk_Panov_Yakovlev_2023} for predicting correction factor and introduce two modifications: long skip connections between matching-resolution encoder and decoder blocks, and an explicitly masked regression loss. Our model is depicted in Fig.~\ref{fig:arch}. First, the input is processed with the convolution-based encoder that is meant to capture the geometric details such as corners, corridor boundaires etc. An encoder-produced  feature map of dimensions $(C, H',W')$ is reshaped into a sequence of $H'W'$ tokens of dimension $C$, is combined with the learned positional embeddings, and then processed by the sequence of $3$ self-attention blocks. The output tokens are reshaped back to a spatial feature map and fed to the decoder. Importantly, the input to the decoder is augmented with the initial encoder features via long skip connections (merging is performed by elementwise addition). Finally, we map the final logits to $(0,1]$ via a rescaled $\tanh$ and clamp away from zero before using \eqref{eq:h_from_cf}.




\paragraph{Loss}
\label{sec:loss}

In training the model we incorporate a loss-masking strategy that excludes obstacle cells and the goal cell from contributing to the prediction error. 
Specifically, Let $O$ be the binary obstacle mask ($1$ for blocked cells, $0$ otherwise), and let $G$ be the singleton mask for the goal cell. Define a composite mask $M = \neg O\ \&\ \neg G$, so that $M(n)=1$  identifies the open, non-goal cells whose $cf$–values should be learned. We train the network using a standard regression loss $L_2$ between predicted $cf(n)$ and the ground-truth $cf^*(n)$), but only for cells with $M(n)=1$.
\[
L_{cf} = \frac{\sum_n M(n) \cdot (cf(n) - cf^*(n))^2}{\sum_n M(n)}
\]

\subsection{Universal Path Finding (UPF) Evaluation Dataset}
\label{sec:upf}

Indeed, it is impossible to exhaustively evaluate a pathfinding solver (either learnable one or a classical one) across all possible problem instances. On the other hand pathfinding in practice spans a broad spectrum of settings -- from cluttered indoor scenes to open outdoor terrains -- featuring obstacles of varying scale and structure, maps that are either synthetic or captured from real environments, and difficulty levels ranging from trivial to highly challenging. Thus to better approximate the real average performance of a planner one needs to create a (relatively) small but representative set of test tasks. To this end we design a deliberately diverse and challenging evaluation suite that stresses generalization across fundamentally different topologies. Specifically, we construct an evaluation dataset with 20{,}000 tasks, evenly split across 10 topologies (2{,}000 tasks per topology) -- see Fig.~\ref{fig:mapgrid}. Our dataset, dubbed UPF (Universal Pathfinding) consists of the:


\begin{figure*}[ht]
\centering
\begin{tabular}{@{\hspace{2pt}}c@{\hspace{2pt}}c@{\hspace{2pt}}c@{\hspace{2pt}}c@{\hspace{2pt}}c}
  \includegraphics[width=0.18\textwidth]{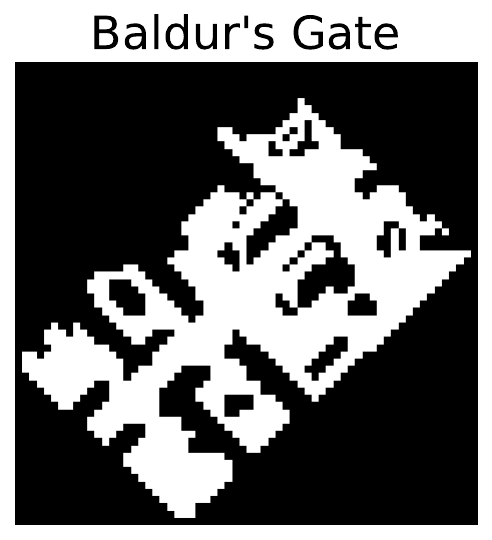} &
  \includegraphics[width=0.18\textwidth]{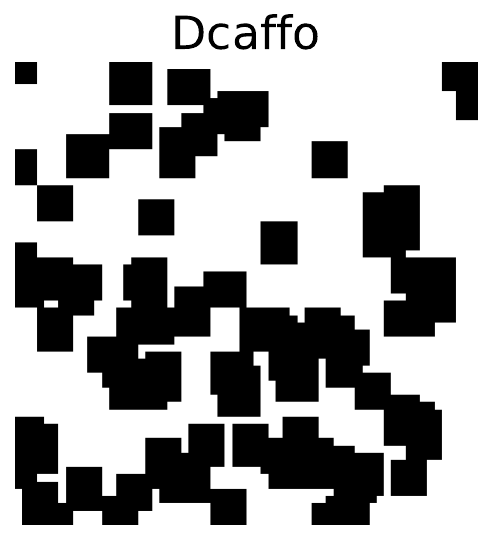} &
  \includegraphics[width=0.18\textwidth]{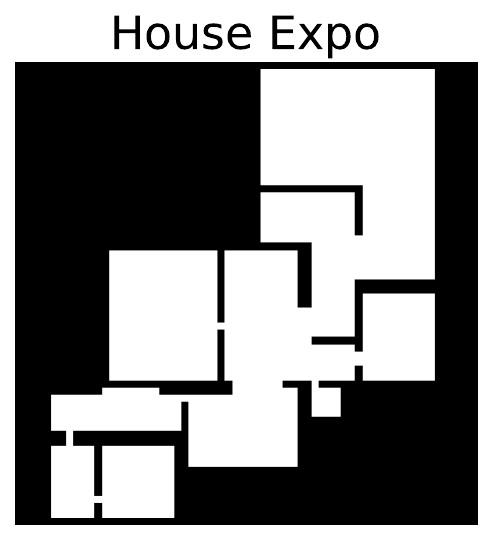} &
  \includegraphics[width=0.18\textwidth]{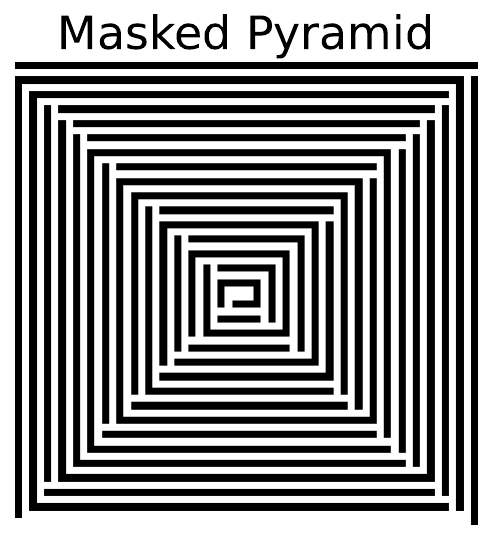} &
  \includegraphics[width=0.18\textwidth]{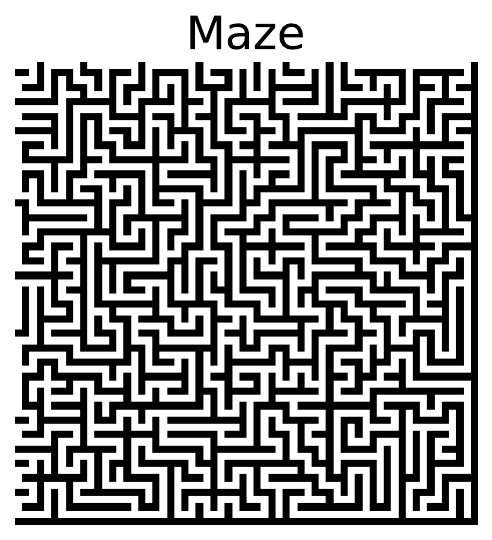} \\
  \includegraphics[width=0.18\textwidth]{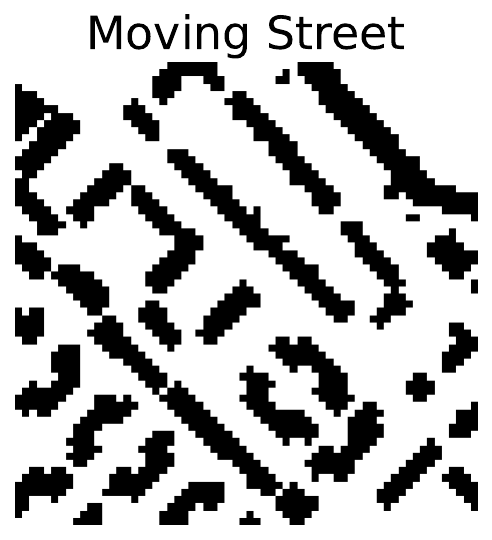} &
  \includegraphics[width=0.18\textwidth]{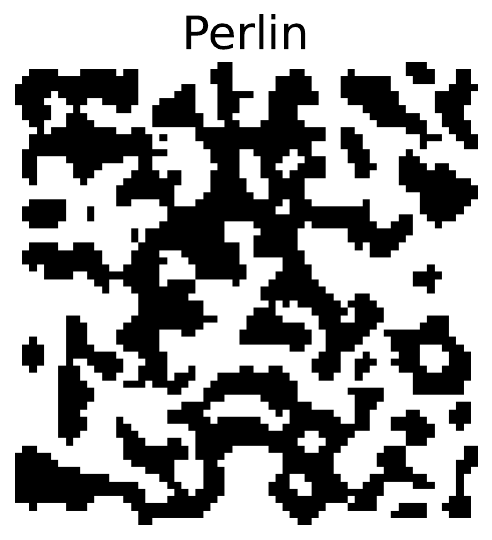} &
  \includegraphics[width=0.18\textwidth]{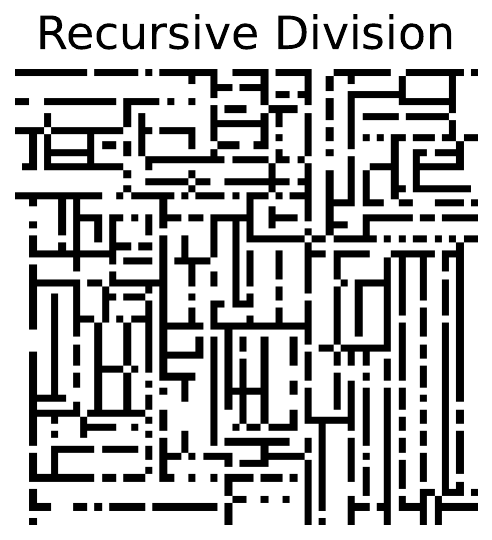} &
  \includegraphics[width=0.18\textwidth]{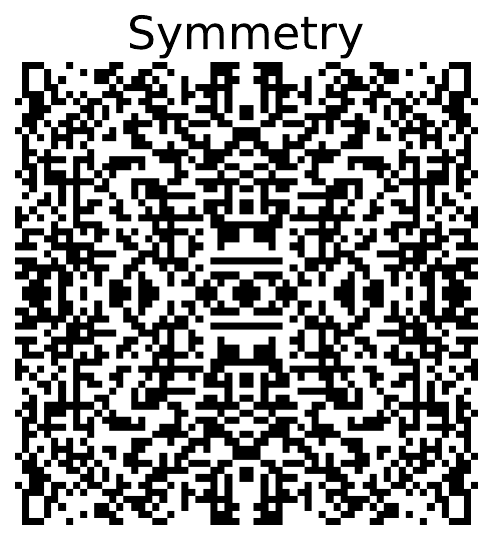} &
  \includegraphics[width=0.18\textwidth]{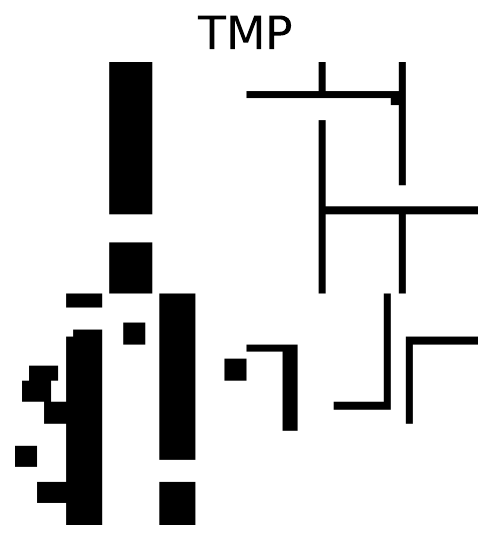} \\
\end{tabular}
\caption{UPF topologies, 64x64.}
\label{fig:mapgrid}
\end{figure*}

\begin{enumerate}
    \item Established sources: Baldur’s Gate and Moving Street \cite{sturtevant2012benchmarks}, TMP \cite{pmlr-v139-yonetani21a}, and HouseExpo (adapted from \cite{Li2019HouseExpo}).
    \item Noise-generated layouts: Perlin, Dcaffo \cite{davide_caffagni_2022}, and our Rotational Symmetry and Recursive Division generators.
    \item Maze-like structures: Prim-style mazes and our Masked Pyramid generator.
\end{enumerate}

Each topology corresponds to a procedural $64\times64$ map generator that produces a binary grid. Additional details on the dataset are provided in the Appendix~\ref{sec:test-dataset-details}.


\paragraph{Task generation and filtering.}
For each generated grid $G$, we sample a random goal cell and a start one that is definitely reachable. To avoid degenerate tasks, we enforce two criteria:



\emph{Reachability Diversity.}
Let $R$ be the set of traversable cells reachable from the goal. Define
\begin{equation}
    H(\textit{goal},G) \;=\; \sum_{s\in R} h^{*}(s),
\end{equation}
a proxy for the size/shape of the reachable component. We require
$H(\textit{goal},G)\ge H_{\min}$. For $64\times64$ grids we set $H_{\min}=553$. This specific value is computed on a fully traversable $11\times11$ grid with the centered goal.

\emph{Reachability Complexity.}
We retain only tasks whose cost satisfies
\begin{equation}
    \mathrm{cost}(\pi^*(start,goal)) \;\ge\; 1.05\cdot \mathrm{h_{oct}}(\textit{start}),
\end{equation}
which removes near-straight-line instances.

\section{Empirical Evaluation}

\subsection{Training}

We train three models, one per training dataset (Uniform, Beta, Beta-Figures). For all models we use the same training protocol. Each model is optimized with Adam \cite{kingma2017adammethodstochasticoptimization} for 50 epochs with batch size 512, and we use a OneCycleLR learning-rate schedule \cite{smith2018superconvergencefasttrainingneural} with a peak learning rate of $8 \times 10^{-3}$. On a single NVIDIA A100 (40GB), training on a dataset of 512000 tasks with input shape $(2,64,64)$ takes approximately 2.5 hours per model.

\subsection{Evaluation Setup}

We evaluate all planners on the UPF benchmark. We compare with the following baselines: Weighted A* (WA*) with $w \in \{2, 5, 10\}$ and state-of-the-art learnable solver, TransPath~\cite{}, which also predicts correction factor and was previoulsy shown to outperform all other learning-based competitors (so we omit including them in our tests). We use the authors' official implementation of TransPath and the released weights.



Our primary metric is the \emph{expansions ratio}, which is the number of nodes expanded by a solver divided by the number of nodes expanded by A* with the octile-distance heuristic. This ratio serves as a proxy for computational effort. To assess solution quality, we measure the \emph{optimal found ratio}, i.e., the fraction of instances in which the returned path is optimal, and the \emph{cost ratio}, i.e., the returned path length divided by the optimal path length. 
Unlike most of the prior work, we also evaluate runtime: for each method we measure the total runtime (including prediction time for learning-based solvers) used to solve all instances from the test dataset.


\subsection{Results}
Table \ref{tab:results64} reports the mean values and standard errors of our metrics in the evaluation set. As the results demonstrate, all of our learning-based planners generalize effectively to previously unseen instances, achieving near-optimal solutions while substantially reducing search effort.

\begin{table}[ht]
\centering
\resizebox{\columnwidth}{!}{
\begin{tabular}{l|ccc}
        & Optimal Found & Cost & Exp \\
        & Ratio ($\%$) $\uparrow$ & Ratio ($\%$) $\downarrow$ & Ratio ($\%$) $\downarrow$ \\
        \hline
        {A*}           & 100.00 & 100.0          & 100.0  \\
        {UPath (Uniform)}           & 63.23 & \textbf{101.1$\pm$2.9}          & 53.8$\pm$29.0  \\
        {UPath (Beta)}         & 55.24  & 105.1$\pm$16.2 & 45.3$\pm$31.9  \\
        {UPath (Beta+Fig)}    & \textbf{72.63}  & \textbf{101.1$\pm$4.1}  & 47.4$\pm$27.7  \\
        WA*, w=2    & 32.35  & 103.7$\pm$4.9  & 54.6$\pm$30.1  \\
        WA*, w=5    & 14.38  & 107.9$\pm$8.9  & 47.3$\pm$29.8  \\
        WA*, w=10    & 13.40  & 109.7$\pm$11.1  & \textbf{45.1$\pm$29.2}  \\
        {TransPath}    & {32.34}  & {125.9}$\pm$49.7  & 111.4$\pm$134.3  \\
\end{tabular}
}
\caption{Performance comparison on 64x64 UPF.}
\label{tab:results64}
\end{table}

Among the learning-based planners, UPath (Beta+Fig) is characterized by the best performance: it attains the highest Optimal Found Ratio (72.63\%) while keeping solution quality essentially near-optimal (101.1$\pm$4.1\%) and still reducing search effort by roughly 2.11$\times$ on average (Expansions Ratio 47.4$\pm$27.7\%). UPath (Uniform) matches the best cost accuracy (101.1$\pm$2.9\%) but is less reliable in optimal solves (63.23\%) and requires noticeably more expansions (53.8$\pm$29.0\%), indicating weaker generalization to harder topologies despite similar path quality. UPath (Beta) is the most aggressive in pruning: it achieves the lowest Expansions Ratio (45.3$\pm$31.9\%), i.e., 2.21$\times$ fewer expansions than A* on average, but this speed comes with a substantial drop in optimal-solve rate (55.24\%) and increased solution cost (105.1$\pm$16.2\%).

The WA* baselines exhibit the expected trend: increasing $w$ improves average expansions (down to 45.1$\pm$29.2\% at $w{=}10$) but sharply degrades optimality (only 13--32\% optimal found) and progressively inflates cost (up to 109.7$\pm$11.1\%), placing them on a dominated part of the frontier relative to UPath variants that achieve similar or better expansion reductions with markedly higher optimal-solve rates.

Finally, TransPath is decisively outperformed on UPF: it expands more nodes than A* on average (111.4$\pm$134.3\%) and incurs the worst cost inflation (125.9$\pm$49.7\%), with very high variance. This sharp degradation relative to its originally reported setting is consistent with a strong sensitivity to the evaluation distribution: when the benchmark departs from the training-like regime, the learned heuristic can become miscalibrated, leading to both excessive search and poor solution quality.

\begin{figure}[t]
    \centering
    \includegraphics[width=\columnwidth]{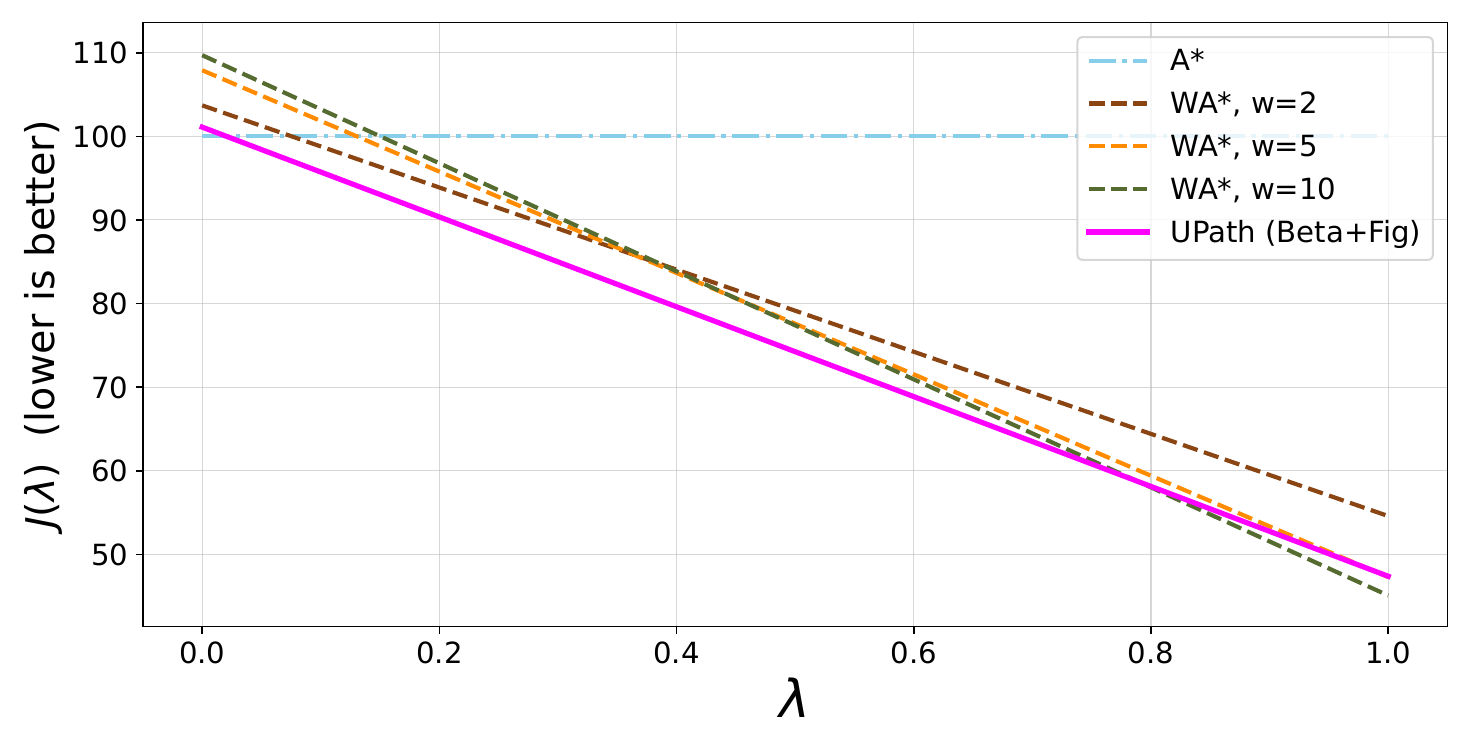}
    \caption{Trade-off analysis.}
    \label{fig:tradeoff}
\end{figure}

Figure~\ref{fig:tradeoff} plots the combined cost-expansion objective $ J(\lambda)$ for $\lambda$ sweeping all the way from quality-dominated regimes ($\lambda\!\rightarrow\!0$) to compute-dominated regimes ($\lambda\!\rightarrow\!1$).
Across a broad range of $\lambda$, UPath achieves a strictly better trade-off than Weighted A* baselines: in particular, UPath (Beta+Fig)
outperforms WA* with $w=10$ for all $\lambda<0.79$, i.e., whenever path quality is not negligible, while retaining near-optimal costs.

\begin{figure}[ht]
    \centering
    \includegraphics[width=\columnwidth]{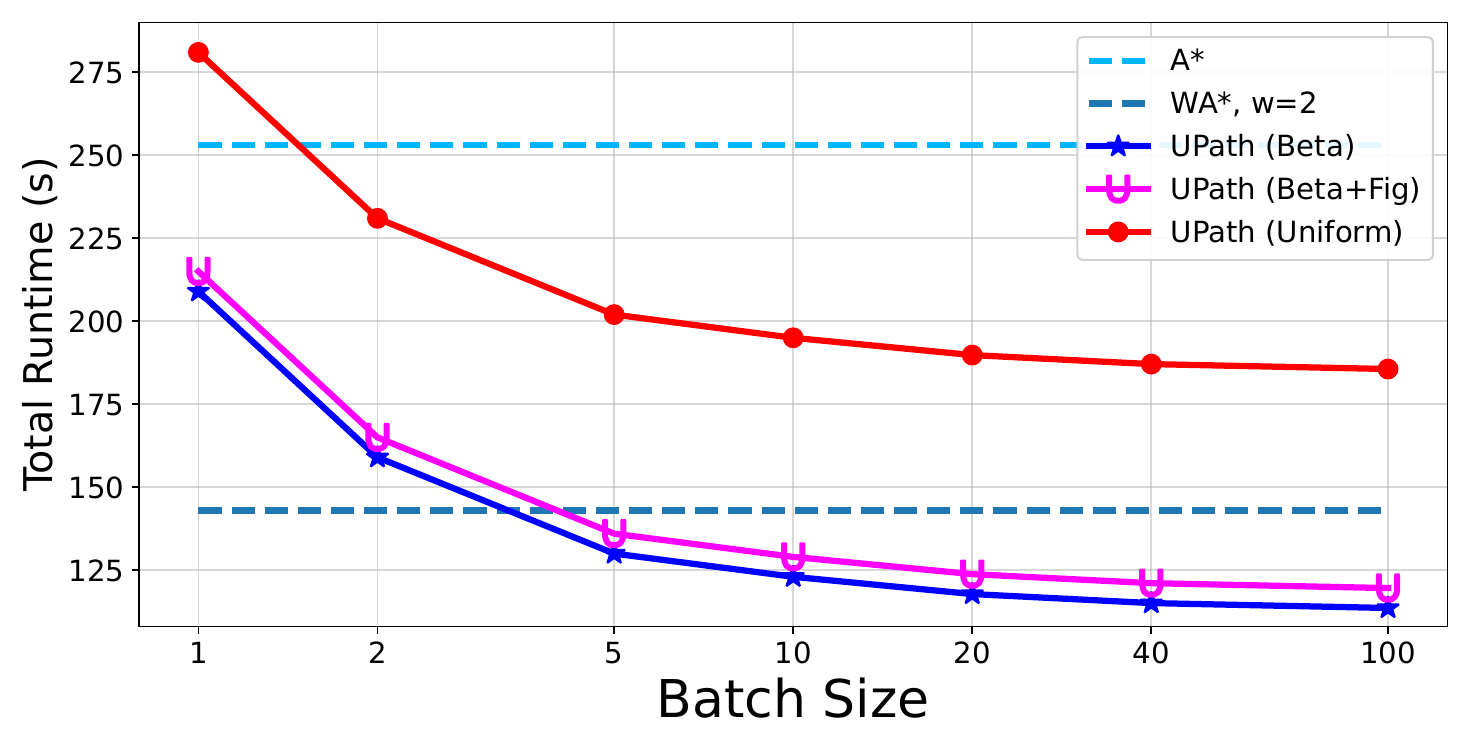}
    \caption{Total runtime (in seconds) as a function of batch size.}
    \label{fig:runtime}
\end{figure}

Figure~\ref{fig:runtime} reports the runtime consumed to solve all the problem instances for best-performing algorithms.
The results indicate that with a batch size of $5$, UPath (Beta+Fig) and UPath (Beta) solve tasks faster than WA* with $w=2$, whereas UPath (Uniform) is slower. This shows that our method delivers exceptional speed even with very small batch sizes.

\paragraph{Ablation.}
We ablate the two components introduced in Section~\ref{sec:model} by training UPath (Beta) with (i) skip connections removed, and (ii) both skip connections and loss masking removed.
Results are summarized in Table~\ref{tab:ablation_flags}.

As one can note, removing the skip connections makes the search consistently less efficient, increasing expansions from $45.3\%$ to $53.65\%$, with a small degradation in the remaining metrics. In contrast, removing loss masking produces a much larger drop: expansions further rise to $77.4\%$ and the success rate collapses (Optimal Found $51.64\%\!\rightarrow\!6.40\%$), with higher path costs as well, indicating that masking is crucial for robust training and transfer.

\newcommand{\cmark}{\ensuremath{\checkmark}}
\newcommand{\xmark}{\ensuremath{\times}}

\begin{table}[ht]
\centering
\resizebox{\columnwidth}{!}{
\begin{tabular}{cc|ccccc}
        Skips & Masking &  Optimal Found & Cost & Expansions \\
              &             & Ratio ($\%$) $\uparrow$ & Ratio ($\%$) $\downarrow$ & Ratio ($\%$) $\downarrow$ \\
        \hline
        \cmark & \cmark & 55.24  & 105.1$\pm$16.2 & 45.3$\pm$31.9 \\
        \xmark & \cmark & 51.64 & 105.7$\pm$16.75 & 53.65$\pm$31.14 \\
        \xmark & \xmark & 6.40 & 110.6$\pm$15.11 & 77.4$\pm$36.2 \\
\end{tabular}
}
\caption{Ablation on long skip connections and loss masking.}
\label{tab:ablation_flags}
\end{table}

\paragraph{Scaling to larger maps.}
To assess scalability, we also train and evaluate UPath on $128 \times 128$ maps. The obtained results demonstrate that the suggested solver performs well at higher resolutions (i.e. outperform competitors similarly to the presented $64 \times 64$ evaluation. The details of this experiment are provided in the Appendix~\ref{sec:additional-experiments}.

\section{Conclusion}
We introduced UPath, a \emph{train-once, search-everywhere} heuristic predictor for grid-based pathfinding under strong topological shift. UPath learns an instance-dependent correction-factor map relative to the octile prior and can be plugged directly into standard A* without modifying the search procedure. To make universality testable, we also proposed UPF, a deliberately topology-diverse evaluation suite covering ten qualitatively different map sources.

Across UPF, UPath consistently improves the efficiency--quality frontier: it reduces node expansions by up to $2.2\times$ while keeping solution cost within $3\%$ of optimal, and it dominates common Weighted A* settings by achieving comparable (or lower) expansion ratios at substantially higher optimal-solve rates. Moreover, compared to the state-of-the-art learned heuristic planner TransPath, UPath remains robust on this topology-diverse benchmark, highlighting the importance of evaluating learned heuristics beyond training-like regimes. Finally, results on $128\times128$ tasks indicate that the approach scales favorably when trained at the target resolution.



\bibliographystyle{named}
\bibliography{ijcai26}

\appendix
\include{appendix.tex}

\end{document}

%% file: appendix.tex

\maketitle

\section{Test Dataset Details}
\label{sec:test-dataset-details}
Below we provide details on the Universal Pathfidning Dataset (UPF) used in our experiments. UPF contains 20,000 different maps, split between 10 types of maps of varying topologies (2,000 maps per type):


\begin{enumerate}
    \item \textbf{Baldur’s Gate \cite{sturtevant2012benchmarks}.}
    We sample random $512\times512$ map from the original dataset (consisting of 75 maps) and produce our map via downsampling to the target resolution.
    To increase diversity, we additionally apply $90^\circ$ rotations (i.e., multiples of $90^\circ$).

    \item \textbf{Dcaffo  \cite{davide_caffagni_2022}.}
    We generate binary noise map and apply a morphological convolution operator with fixed parameters (as in the referenced implementation).
    In our codebase we additionally provide parameter settings for larger map sizes ($256\times256$ and $512\times512$), which were not included in the original configuration.

    \item \textbf{HouseExpo (adapted from \cite{Li2019HouseExpo}).}
    We sample a random floorplan from the full pool of 35126 maps from the original dataset. Each sampled map is converted to the target resolution using a custom resizing procedure: depending on the source size, we either downsample the map or pad the boundary region with obstacles (non-traversable cells) to match the desired dimensions.

    \item \textbf{Masked Pyramid (ours).}
    We construct a pyramid consisting of 15 concentric layers. Layers are contiguous (solid) except for corner cells at each layer. For every layer, we designate four corner-related cells whose traversability is randomized as follows:
    with probability $0.25$ we mark exactly 1 of the 4 cells as blocked; with probability $0.25$ we mark exactly 3 cells as blocked; and with probability $0.5$ we mark exactly 2 cells as blocked.

    \item \textbf{Prim Maze (maze-like generator).}
    We initialize a fully blocked grid, select an initial traversable cell, and add its blocked neighbors to a frontier list. While the frontier list is non-empty, we pop a cell from the list; if it has exactly one traversable neighbor, we mark it traversable and add its blocked neighbors to the frontier. The processed cell is then removed from the list.
    This procedure produces maze-like corridor structures.

    \item \textbf{Moving Street \cite{sturtevant2012benchmarks}.}
    We utilize the original Moving AI maps, that are $4\times$ larger (per side) than our target resolution.
    We first crop a random sub-window of size $2\times$ the target side length, and then downsample by a factor of 2 (per side) to obtain the final map.

    \item \textbf{Perlin (ours)}
    We sample an i.i.d.\ binary noise grid and apply two iterations of majority-rule smoothing using the $3\times3$ Moore neighborhood:
    a cell becomes blocked if more than 4 cells in its neighborhood (including itself) are blocked; otherwise it becomes traversable.

    \item \textbf{Recursive Division (ours).}
    We recursively partition the grid by alternating horizontal and vertical splits, drawing wall segments along each split.
    To introduce stochastic openings, each wall cell is flipped to traversable with probability $0.2$.

    \item \textbf{Rotational Symmetry noise (ours).}
    We generate noise on one quadrant (one quarter of the map) and then mirror it to the remaining quadrants to enforce rotational symmetry.
    This yields maps whose global structure is easier to exploit than fully unconstrained noise at the same resolution, providing a targeted test of whether models leverage such regularities.

    \item \textbf{TMP \cite{pmlr-v139-yonetani21a}.}
    We use the standard TMP dataset generation pipeline provided by the original authors, without modification.


% ...

\begin{table*}[p]
\centering
\small
\setlength{\tabcolsep}{2pt}
\renewcommand{\arraystretch}{1.1}

\begin{tabular}{@{} >{\raggedright\arraybackslash}m{0.18\textwidth} c c c c c @{}}
\toprule
\textbf{UPF Topology} & \textbf{S1} & \textbf{S2} & \textbf{S3} & \textbf{S4} & \textbf{S5} \\
\midrule
Baldur's Gate
& \includegraphics[width=0.12\linewidth]{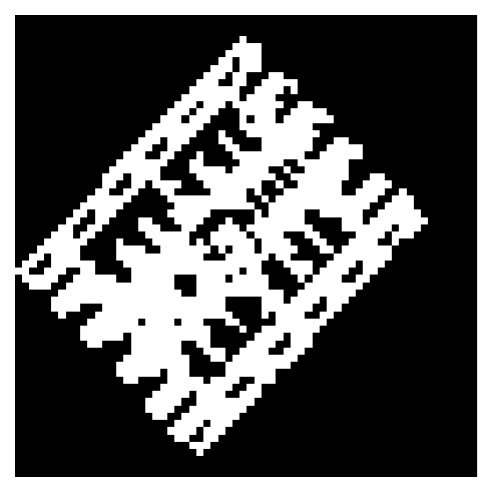}
& \includegraphics[width=0.12\linewidth]{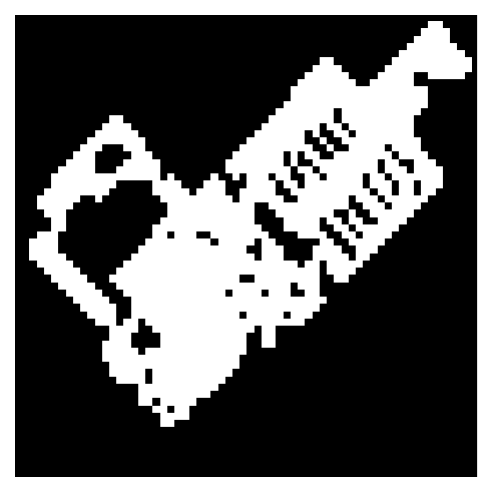}
& \includegraphics[width=0.12\linewidth]{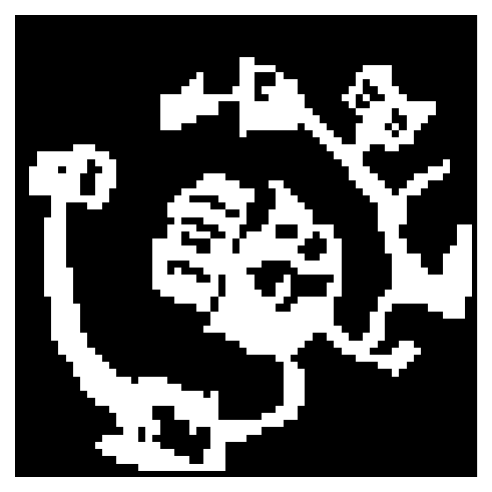}
& \includegraphics[width=0.12\linewidth]{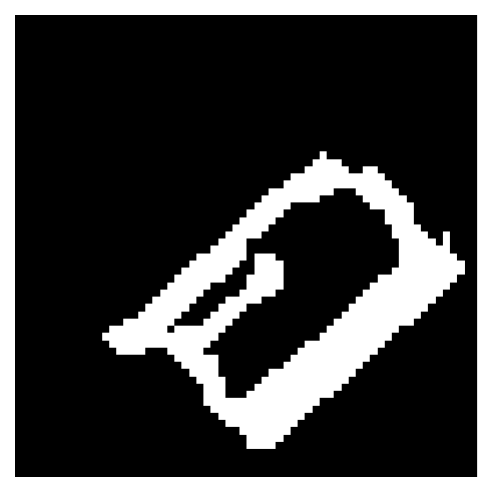}
& \includegraphics[width=0.12\linewidth]{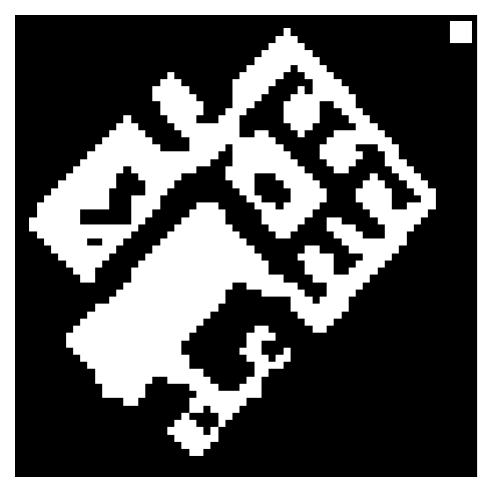} \\

Dcaffo
& \includegraphics[width=0.12\linewidth]{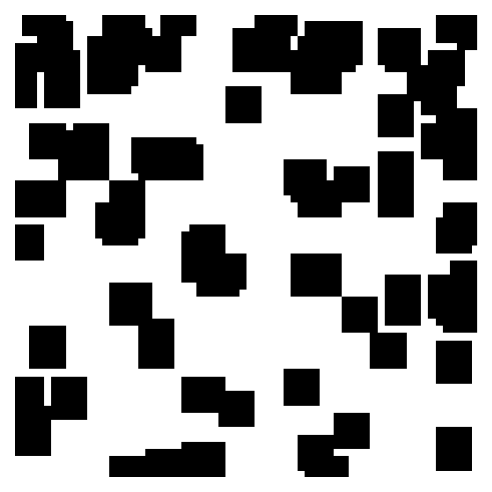}
& \includegraphics[width=0.12\linewidth]{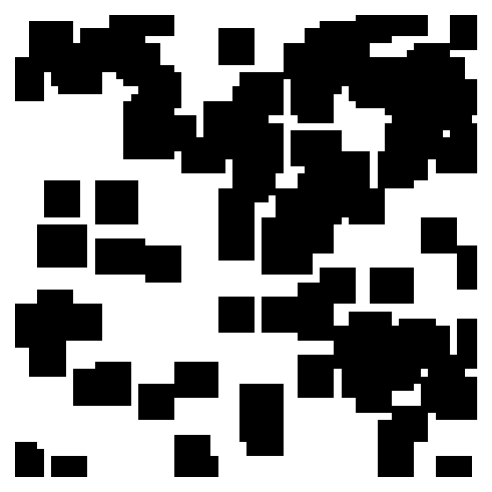}
& \includegraphics[width=0.12\linewidth]{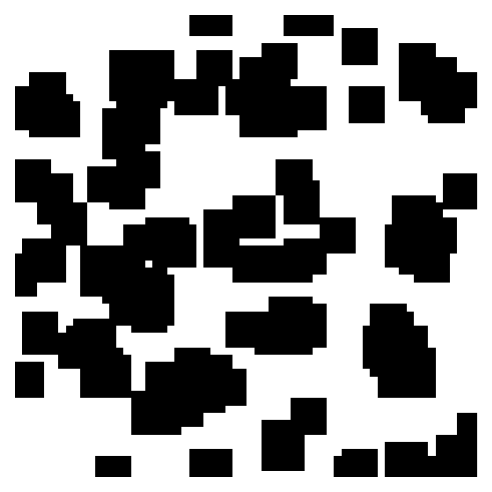}
& \includegraphics[width=0.12\linewidth]{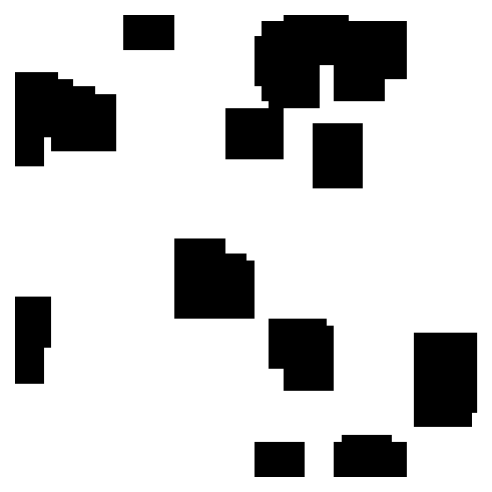}
& \includegraphics[width=0.12\linewidth]{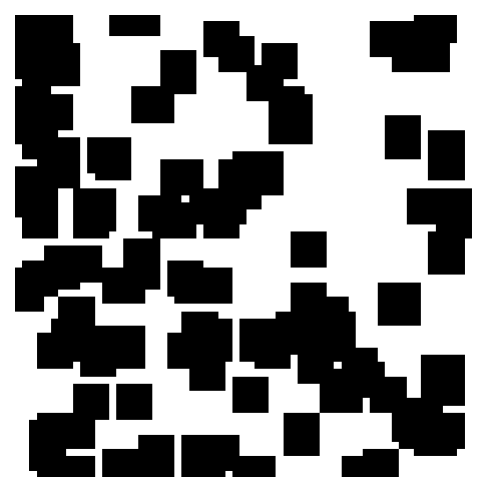} \\

House Expo
& \includegraphics[width=0.12\linewidth]{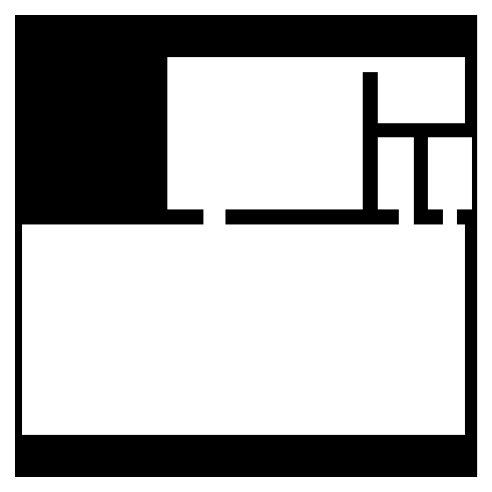}
& \includegraphics[width=0.12\linewidth]{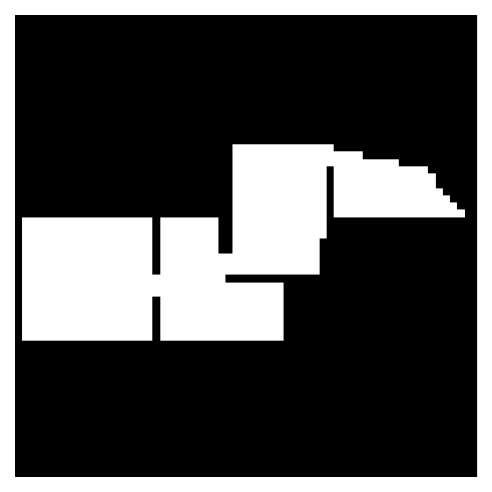}
& \includegraphics[width=0.12\linewidth]{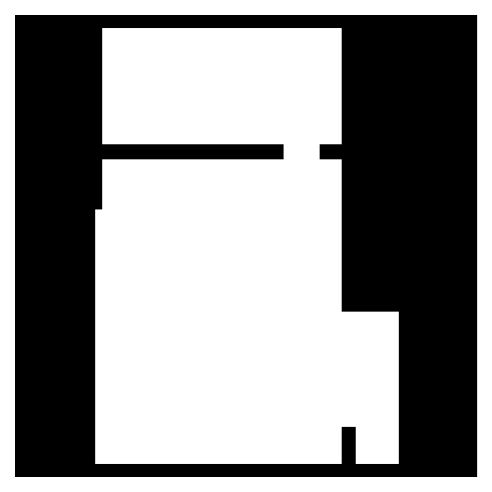}
& \includegraphics[width=0.12\linewidth]{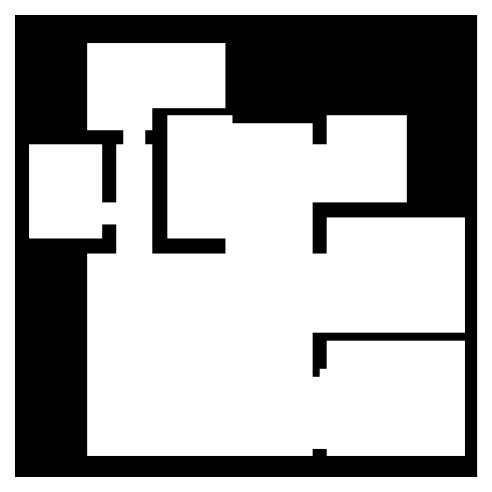}
& \includegraphics[width=0.12\linewidth]{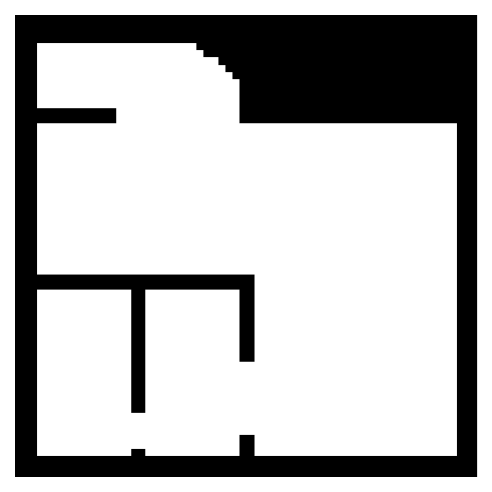} \\

Masked Pyramid
& \includegraphics[width=0.12\linewidth]{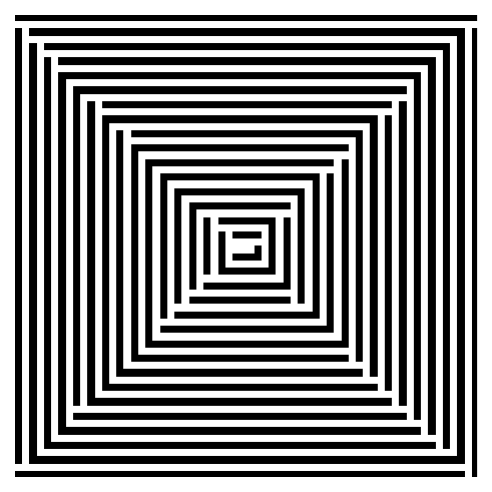}
& \includegraphics[width=0.12\linewidth]{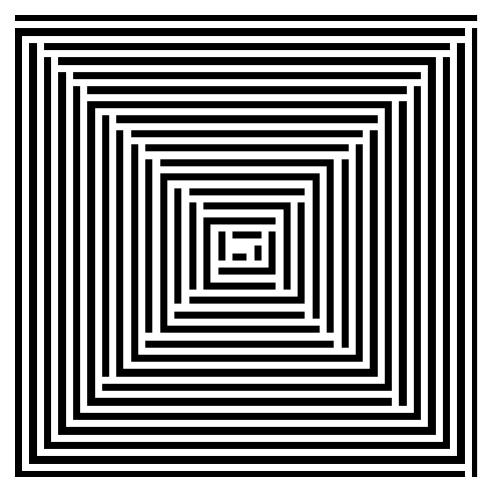}
& \includegraphics[width=0.12\linewidth]{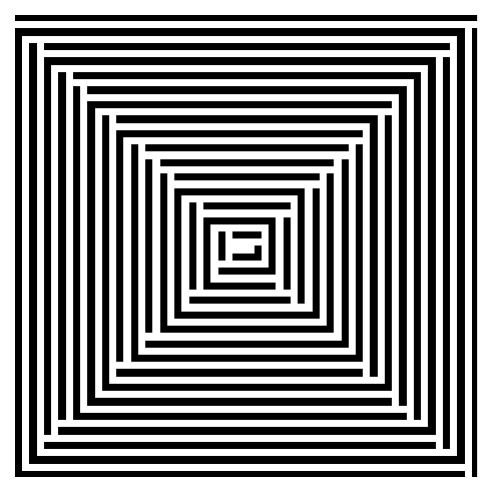}
& \includegraphics[width=0.12\linewidth]{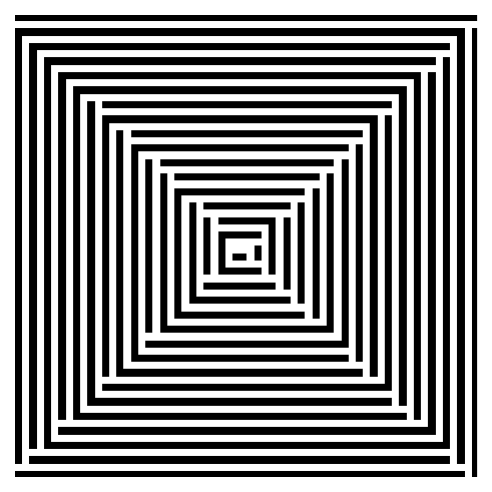}
& \includegraphics[width=0.12\linewidth]{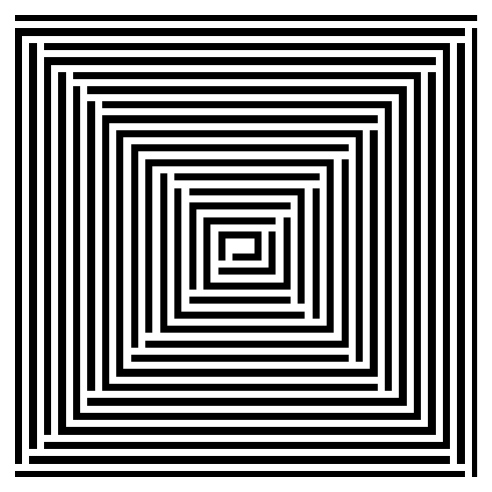} \\

Prim maze
& \includegraphics[width=0.12\linewidth]{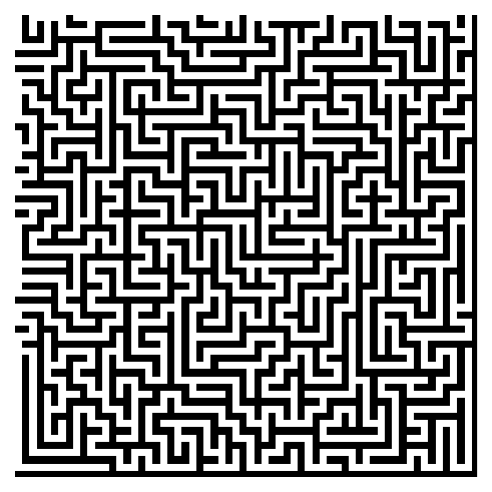}
& \includegraphics[width=0.12\linewidth]{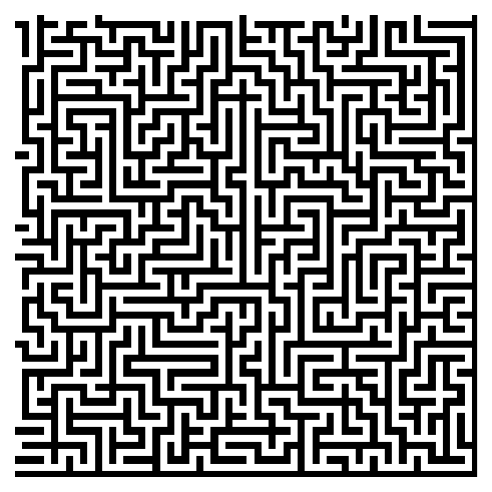}
& \includegraphics[width=0.12\linewidth]{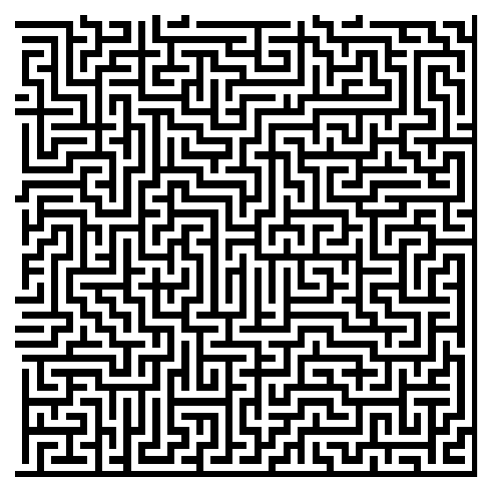}
& \includegraphics[width=0.12\linewidth]{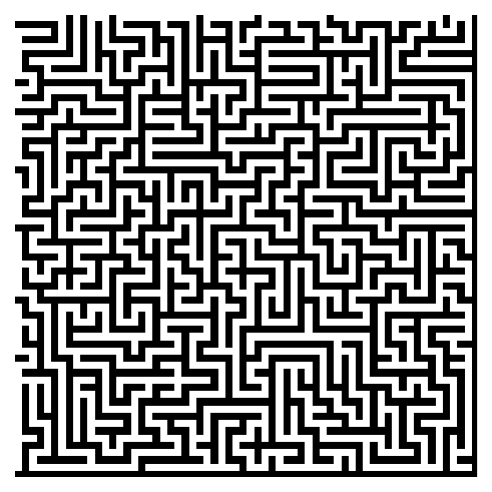}
& \includegraphics[width=0.12\linewidth]{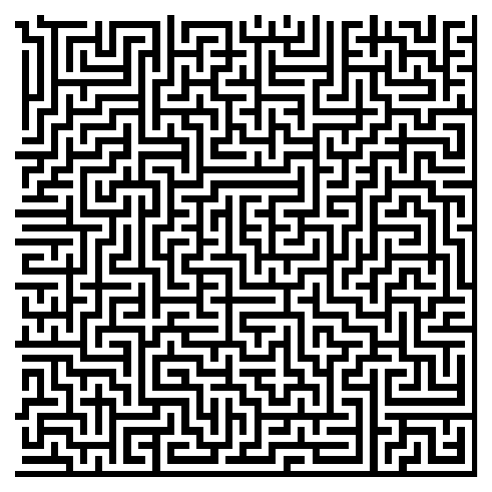} \\

Moving Street
& \includegraphics[width=0.12\linewidth]{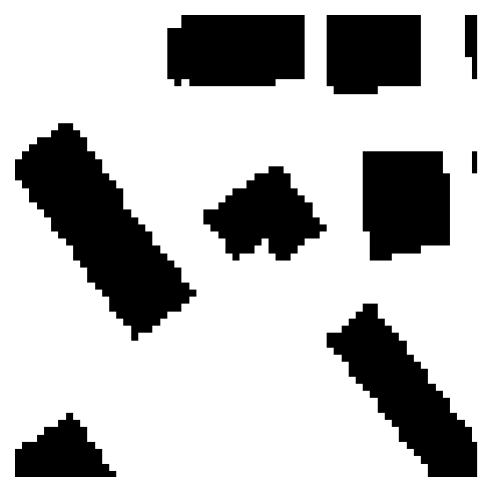}
& \includegraphics[width=0.12\linewidth]{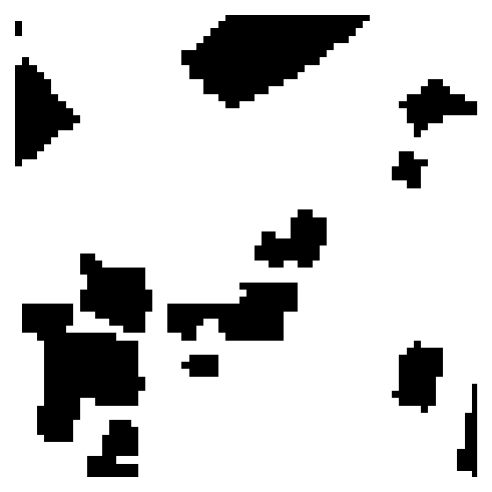}
& \includegraphics[width=0.12\linewidth]{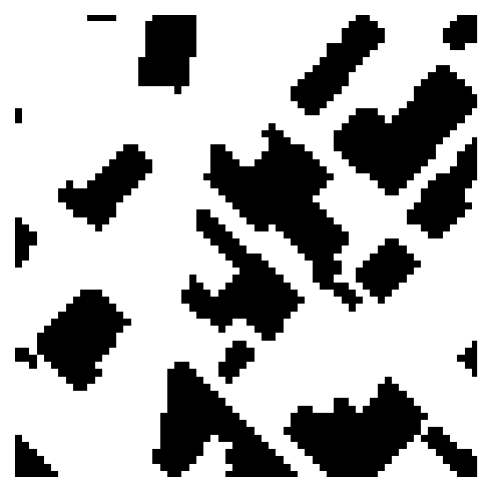}
& \includegraphics[width=0.12\linewidth]{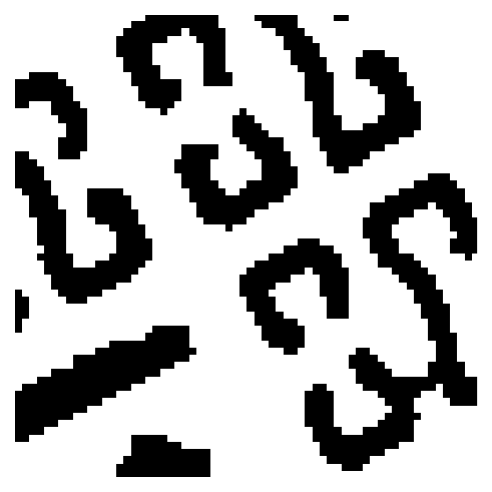}
& \includegraphics[width=0.12\linewidth]{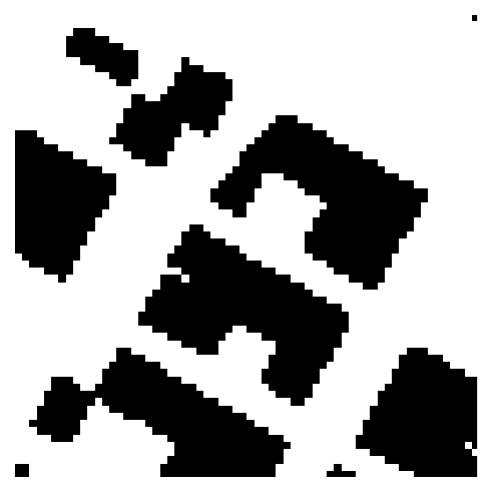} \\

Perlin
& \includegraphics[width=0.12\linewidth]{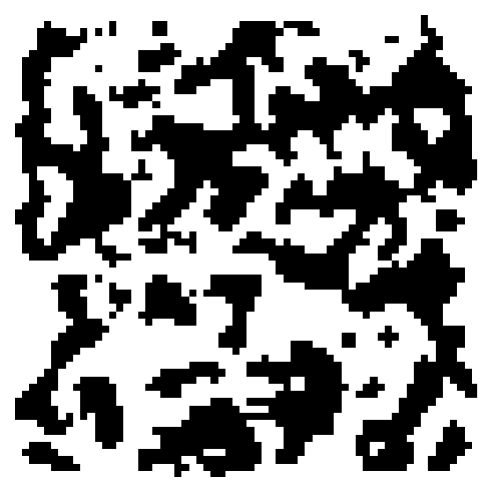}
& \includegraphics[width=0.12\linewidth]{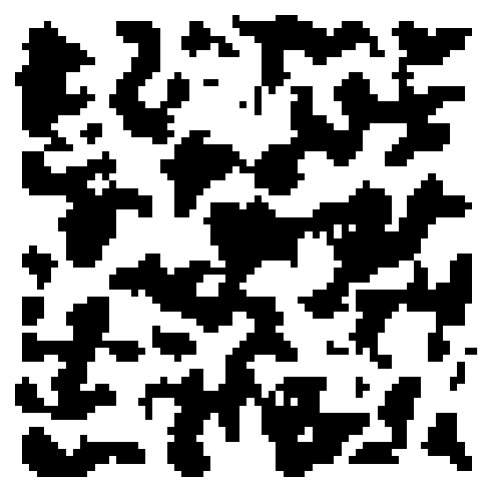}
& \includegraphics[width=0.12\linewidth]{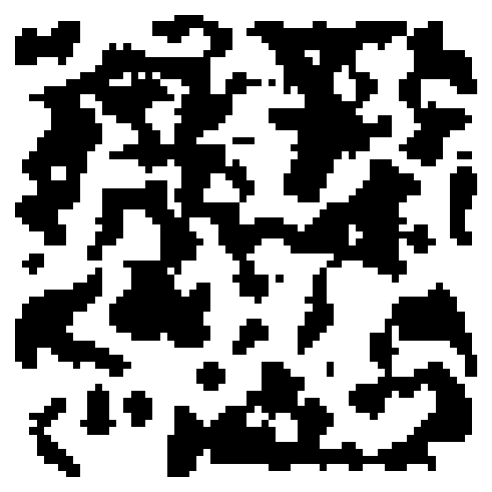}
& \includegraphics[width=0.12\linewidth]{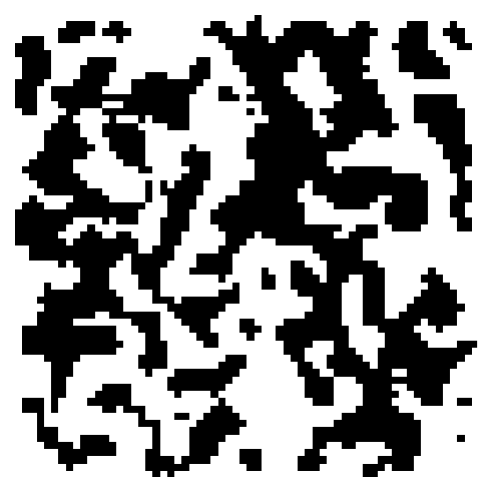}
& \includegraphics[width=0.12\linewidth]{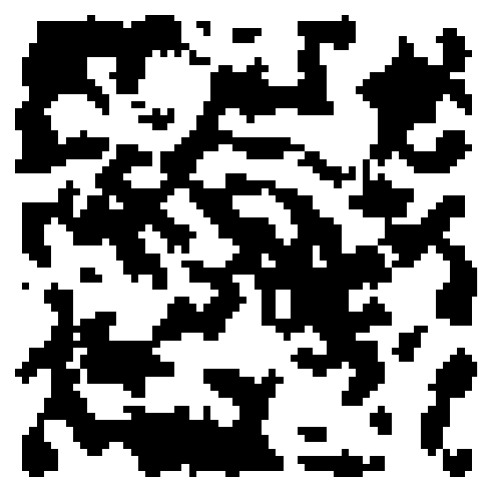} \\

Recursive Division
& \includegraphics[width=0.12\linewidth]{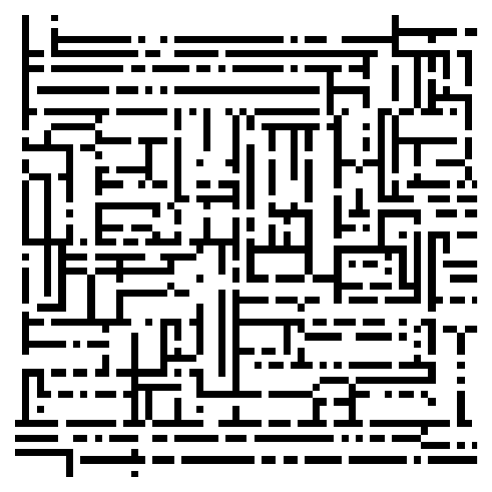}
& \includegraphics[width=0.12\linewidth]{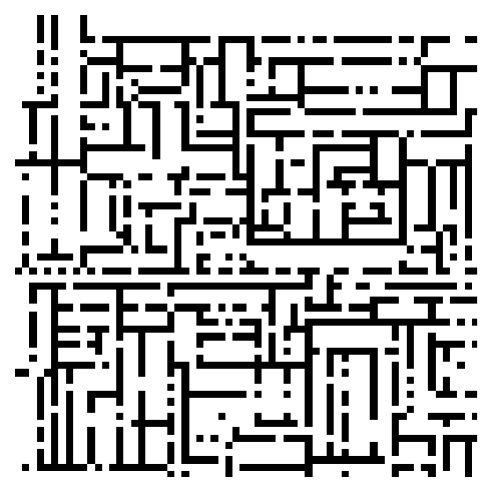}
& \includegraphics[width=0.12\linewidth]{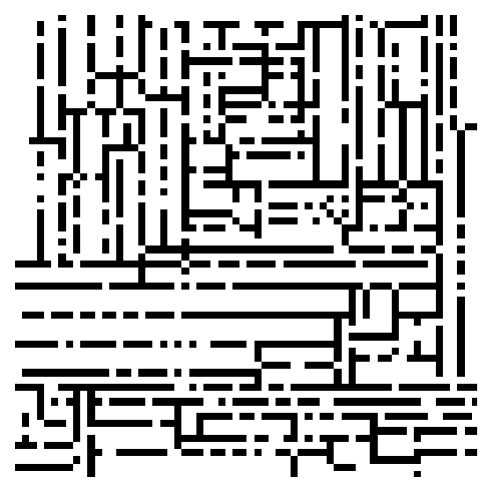}
& \includegraphics[width=0.12\linewidth]{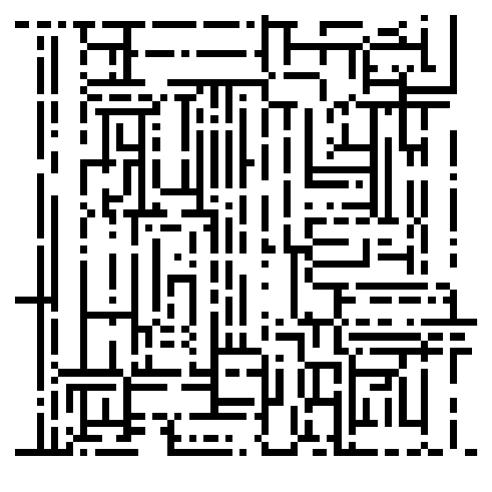}
& \includegraphics[width=0.12\linewidth]{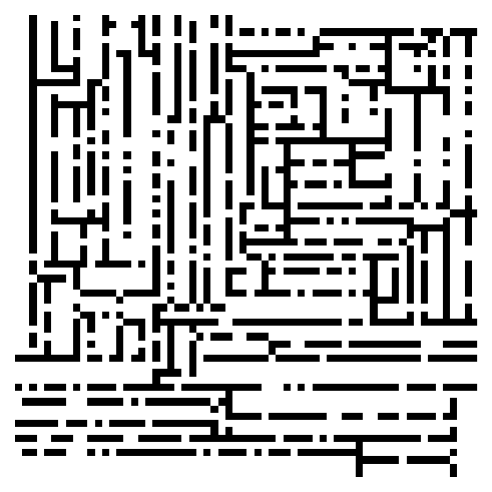} \\

Rotational Symmetry
& \includegraphics[width=0.12\linewidth]{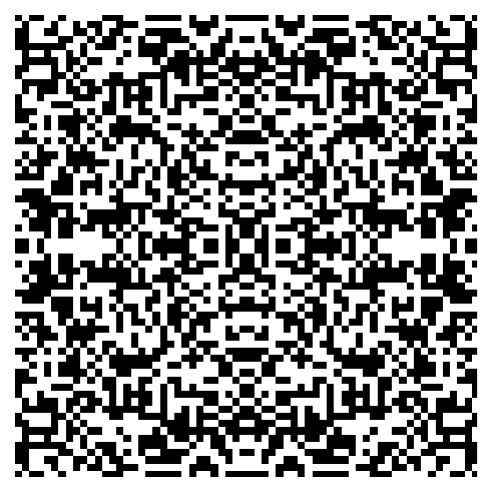}
& \includegraphics[width=0.12\linewidth]{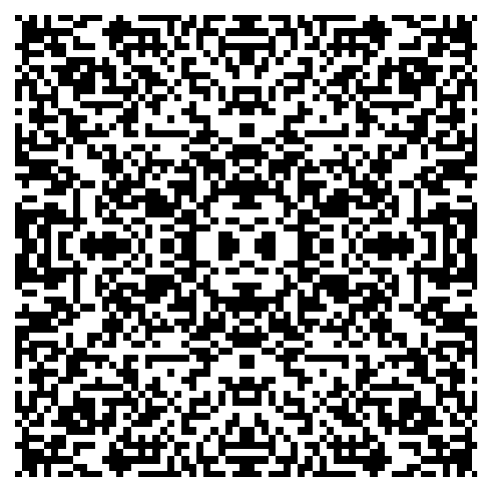}
& \includegraphics[width=0.12\linewidth]{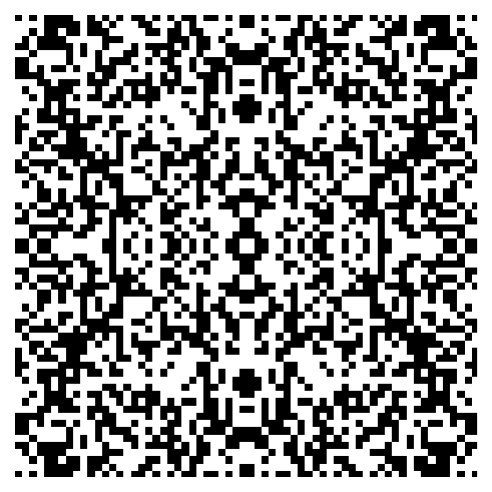}
& \includegraphics[width=0.12\linewidth]{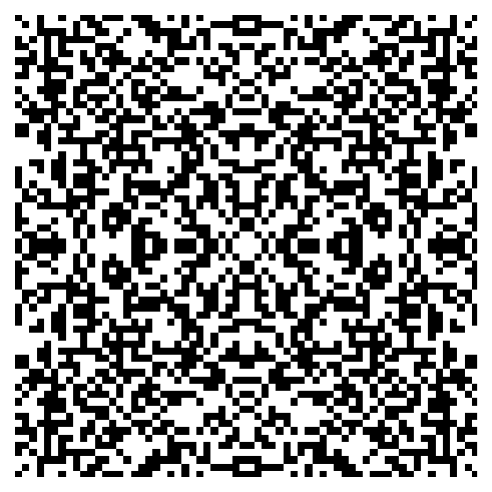}
& \includegraphics[width=0.12\linewidth]{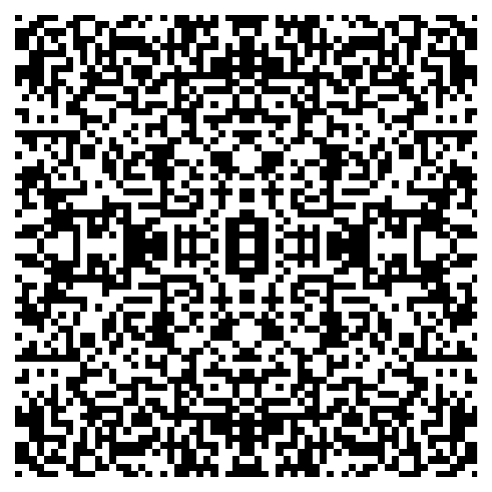} \\

TMP
& \includegraphics[width=0.12\linewidth]{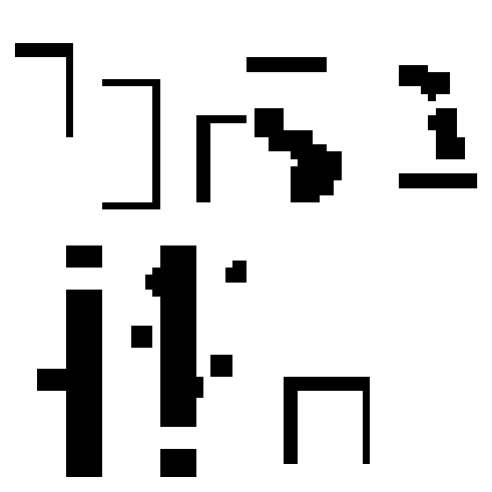}
& \includegraphics[width=0.12\linewidth]{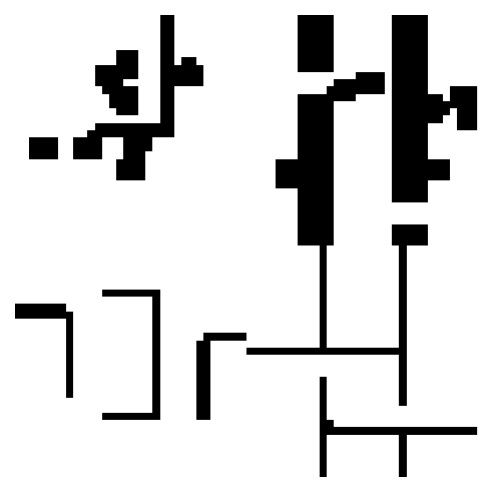}
& \includegraphics[width=0.12\linewidth]{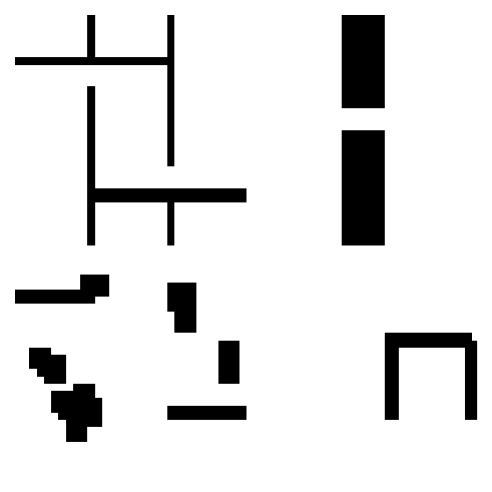}
& \includegraphics[width=0.12\linewidth]{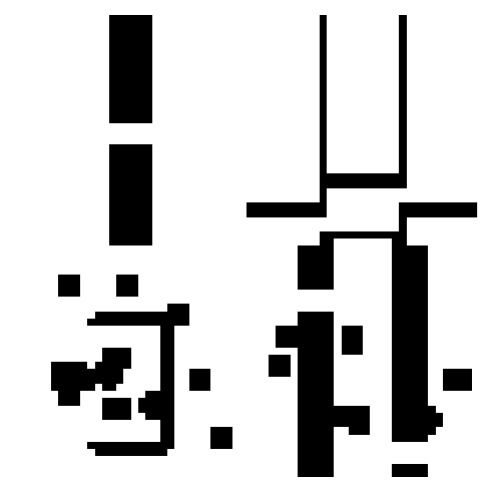}
& \includegraphics[width=0.12\linewidth]{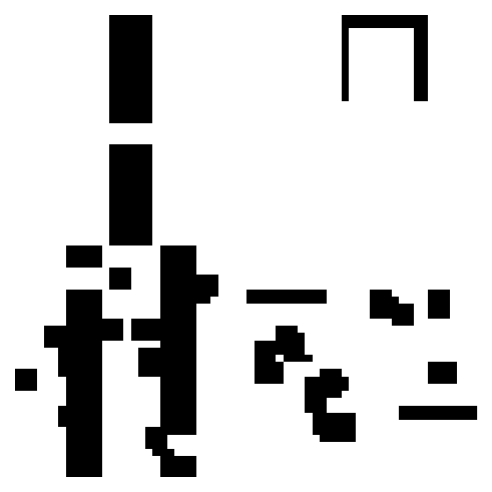} \\
\bottomrule
\end{tabular}

\caption{Representative map samples (five per source/generator).}
\label{tab:dataset_samples}
\end{table*}

\end{enumerate}

\newpage

\section{Additional Evaluation Results}
\label{sec:additional-experiments}

Below we report results of the additional experiments involving maps of larger size (compared to the main experiment), i.e. $128 \times 128$.

\subsection{Training}

For the $128 \times 128$ setting, we train the model on the Beta-Figures dataset, since it achieved the best cost--runtime trade-off on $64 \times 64$ maps. Training follows the same recipe as in the main experiments: we optimize with Adam \cite{kingma2017adammethodstochasticoptimization} for 50 epochs using batch size 512, and apply a OneCycleLR schedule \cite{smith2018superconvergencefasttrainingneural} with peak learning rate $8 \times 10^{-3}$. On a single NVIDIA A100 40GB GPU, training on 512000 tasks with input shape $(2,128,128)$ takes approximately 8.5 hours per model. The increased runtime is primarily due to using a 4-layer encoder--decoder, rather than 3 layers, to better handle the higher input resolution.

\subsection{Evaluation Setup}

We evaluate all planners on the UPF benchmark with $128 \times 128$ maps, using the same WA* baselines as in the $64 \times 64$ experiments. We do not evaluate TransPath in this setting, since its published implementation and released artifacts are built around $64 \times 64$ grid maps, and the performance of the original TransPath on $128 \times 128$ maps is not satisfactory (as we have evidenced after a series of preliminary experiments). 
In addition, to probe scale generalization, we apply UPath trained on Beta-Figures at $64 \times 64$ directly to $128 \times 128$ tasks, without any fine tuning.

We use the same evaluation metrics as in the $64 \times 64$ setting, namely \emph{expansions ratio}, \emph{optimal found ratio}, and \emph{length ratio}.

\subsection{Results}

We denote the model trained on Beta-Figures at $64 \times 64$ as UPath ($64\times64$) and the model trained on Beta-Figures at $128 \times 128$ as UPath ($128\times128$). Table \ref{tab:results128} presents the mean values and standard errors of the performance metrics on the $128\times128$ evaluation set. As the results demonstrate, all of our learning-based planners generalize effectively to previously unseen instances, achieving near-optimal solutions while substantially reducing search effort. UPath ($128\times128$) performs very well in all three key metrics: optimal-found ratio, average path length, and total expansions. It falls only marginally behind WA* ($w=5$) in expansion count. However, WA* ($w=5$) trades that small efficiency gain for a very steep loss in solution quality: it achieves just $7\%$ optimal-found ratio with paths about $9\%$ longer than optimal — much worse than $\approx54\%$ optimal of UPath ($128\times128$) and $\approx2.4\%$ length inflation, although its expansion ratio is only slightly lower at $41.4\%$ versus A*+UPath’s $44.0\%$. This highlights the classic weighted A* trade‑off: higher w reduces expansions but leads to worse path quality. For UPath trained at $64 \times 64$, performance is weaker than for the $128 \times 128$ variant, but the model still generalizes well and remains robust to larger instances.

\begin{table}[ht]
\centering
\resizebox{\columnwidth}{!}{
\begin{tabular}{l|ccc}
        & Optimal Found & Length & Expansions \\
        & Ratio ($\%$) $\uparrow$ & Ratio ($\%$) $\downarrow$ & Ratio ($\%$) $\downarrow$ \\
        \hline
        A*           & 100.00 & 100.0          & 100.0  \\
        UPath (64x64) & 23.84  & 103.7$\pm$7.0  & 48.79$\pm$31.3  \\
        UPath (128x128) & \textbf{54.20}  & \textbf{102.4$\pm$6.9}  & 44.0$\pm$28.7  \\

        WA*, w=2    & 21.22  & 104.5$\pm$5.3   & 48.2$\pm$33.8  \\
        WA*, w=5    & 7.17   & 109.2$\pm$9.2   & 41.4$\pm$33.4  \\
        WA*, w=10    & 6.44   & 111.02$\pm$11.0   & \textbf{39.3$\pm$32.3}
\end{tabular}
}
\caption{Performance of the evaluated planners on $128 \times 128$ UPF maps.}
\label{tab:results128}
\end{table} 

\paragraph{Runtime breakdown}

\begin{figure}[ht]
    \centering
    \includegraphics[width=\columnwidth]{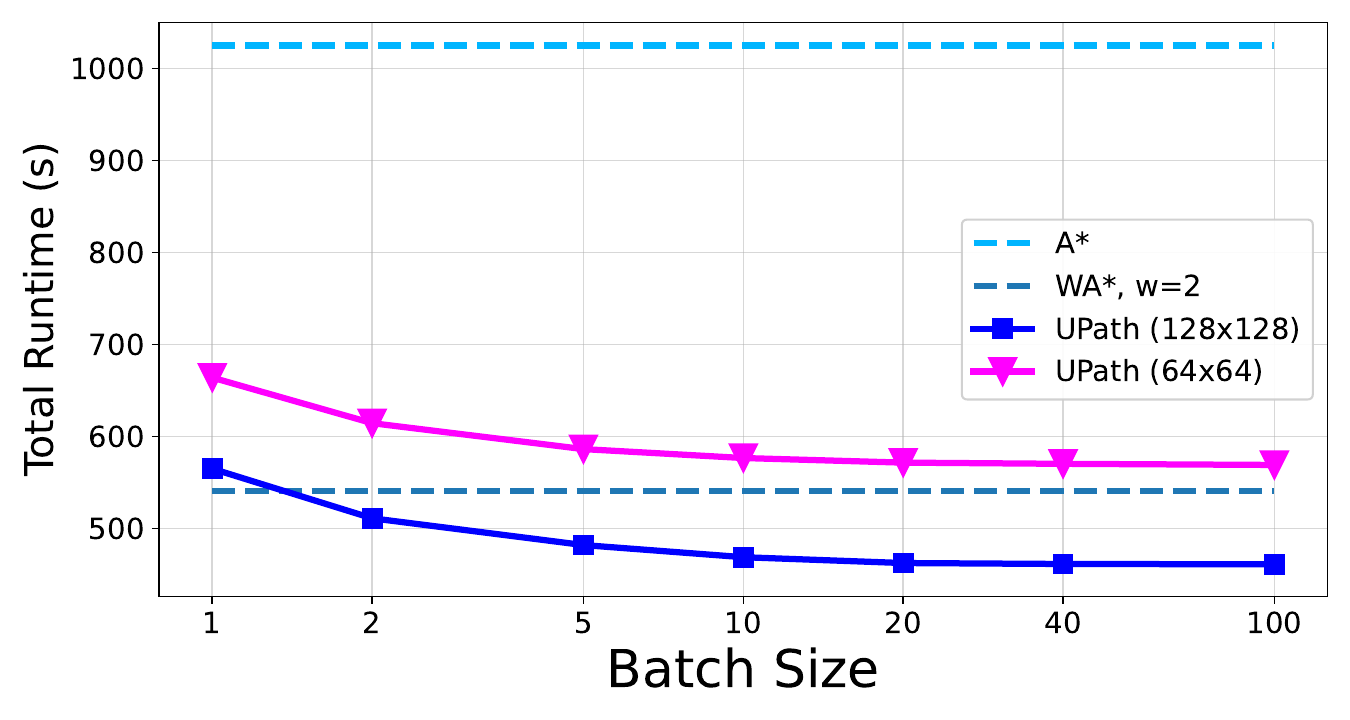}
    \caption{Total runtime (in seconds) on UPF $128\times128$ as a function of batch size.}
    \label{fig:runtime128}
\end{figure}

We evaluated the runtime of each method. Figure \ref{fig:runtime128} reports the total runtime consumed by a solver while solving all instances in the test dataset (the lower the better). X-axis shows the size of the batch used by learnable solvers. For the latter we count both the heuristic prediction time (GPU time) and search time (CPU). Indeed, the prediction time is 0 for A* and WA*. 
The results indicate that when a batch size is greater than $1$, UPath ($128\times128$) solves tasks faster than A* and WA*. And the difference is getting higher with the increase of the batch. For example, when the batch size is 100, UPath ($128\times128$) is approximately 2.22 times faster than A* and 1.17 times faster than WA*($w=2$).


%% file: ijcai26.bib
@article{Li2019HouseExpo,
  author  = {Tingguang Li and Danny Ho and Chenming Li and Delong Zhu and Chaoqun Wang and Max Q.-H. Meng},
  title   = {HouseExpo: A Large-scale 2D Indoor Layout Dataset for Learning-based Algorithms on Mobile Robots},
  journal = {CoRR},
  volume  = {abs/1903.09845},
  year    = {2019},
  note    = {arXiv:1903.09845}
}

@misc{davide_caffagni_2022,
	title={2d-path-planning-dataset},
	url={https://www.kaggle.com/dsv/3575462},
	DOI={10.34740/KAGGLE/DSV/3575462},
	publisher={Kaggle},
	author={Davide Caffagni},
	year={2022}
}

@article{song2016ssc,
  author     = {Song, Shuran and Yu, Fisher  and Zeng, Andy and Chang, Angel X and Savva, Manolis and Funkhouser, Thomas},
  title      = {Semantic Scene Completion from a Single Depth Image},
  journal 	 = {Proceedings of 30th IEEE Conference on Computer Vision and Pattern Recognition},
  year       = {2017},
}

@inproceedings{pmlr-v139-yonetani21a,
  title     = {Path Planning using Neural {A*} Search},
  author    = {Yonetani, Ryo and Taniai, Tatsunori and Barekatain, Mohammadamin and Nishimura, Mai and Kanezaki, Asako},
  booktitle = {Proceedings of the 38th International Conference on Machine Learning},
  series    = {Proceedings of Machine Learning Research},
  volume    = {139},
  pages     = {12029--12039},
  year      = {2021},
  month     = {18--24 Jul},
  publisher = {PMLR},
  url       = {http://proceedings.mlr.press/v139/yonetani21a.html}
}

@InProceedings{pmlr-v78-bhardwaj17a,
  title = 	 {Learning Heuristic Search via Imitation},
  author = 	 {Bhardwaj, Mohak and Choudhury, Sanjiban and Scherer, Sebastian},
  booktitle = 	 {Proceedings of the 1st Annual Conference on Robot Learning},
  pages = 	 {271--280},
  year = 	 {2017},
  editor = 	 {Levine, Sergey and Vanhoucke, Vincent and Goldberg, Ken},
  volume = 	 {78},
  series = 	 {Proceedings of Machine Learning Research},
  month = 	 {13--15 Nov},
  publisher =    {PMLR},
  url = 	 {https://proceedings.mlr.press/v78/bhardwaj17a.html}
}

@article{sturtevant2012benchmarks,
  title={Benchmarks for Grid-Based Pathfinding},
  author={Sturtevant, N.},
  journal={Transactions on Computational Intelligence and AI in Games},
  volume={4},
  number={2},
  pages={144 -- 148},
  year={2012},
  url = {http://web.cs.du.edu/~sturtevant/papers/benchmarks.pdf},
}

@article{Hart1968,
  author  = {Peter E. Hart and Nils J. Nilsson and Bertram Raphael},
  title   = {A Formal Basis for the Heuristic Determination of Minimum Cost Paths},
  journal = {IEEE Transactions on Systems Science and Cybernetics},
  volume  = {4},
  number  = {2},
  pages   = {100--107},
  year    = {1968},
  doi     = {10.1109/TSSC.1968.300136}
}

@article{Kirilenko_Andreychuk_Panov_Yakovlev_2023, title={TransPath: Learning Heuristics for Grid-Based Pathfinding via Transformers}, volume={37}, url={https://ojs.aaai.org/index.php/AAAI/article/view/26465}, DOI={10.1609/aaai.v37i10.26465}, abstractNote={Heuristic search algorithms, e.g. A*, are the commonly used tools for pathfinding on grids, i.e. graphs of regular structure that are widely employed to represent environments in robotics, video games, etc. Instance-independent heuristics for grid graphs, e.g. Manhattan distance, do not take the obstacles into account, and thus the search led by such heuristics performs poorly in obstacle-rich environments. To this end, we suggest learning the instance-dependent heuristic proxies that are supposed to notably increase the efficiency of the search. The first heuristic proxy we suggest to learn is the correction factor, i.e. the ratio between the instance-independent cost-to-go estimate and the perfect one (computed offline at the training phase). Unlike learning the absolute values of the cost-to-go heuristic function, which was known before, learning the correction factor utilizes the knowledge of the instance-independent heuristic. The second heuristic proxy is the path probability, which indicates how likely the grid cell is lying on the shortest path. This heuristic can be employed in the Focal Search framework as the secondary heuristic, allowing us to preserve the guarantees on the bounded sub-optimality of the solution. We learn both suggested heuristics in a supervised fashion with the state-of-the-art neural networks containing attention blocks (transformers). We conduct a thorough empirical evaluation on a comprehensive dataset of planning tasks, showing that the suggested techniques i) reduce the computational effort of the A* up to a factor of 4x while producing the solutions, whose costs exceed those of the optimal solutions by less than 0.3% on average; ii) outperform the competitors, which include the conventional techniques from the heuristic search, i.e. weighted A*, as well as the state-of-the-art learnable planners. The project web-page is: https://airi-institute.github.io/TransPath/.}, number={10}, journal={Proceedings of the AAAI Conference on Artificial Intelligence}, author={Kirilenko, Daniil and Andreychuk, Anton and Panov, Aleksandr and Yakovlev, Konstantin}, year={2023}, month={Jun.}, pages={12436-12443} }

@misc{kingma2017adammethodstochasticoptimization,
      title={Adam: A Method for Stochastic Optimization}, 
      author={Diederik P. Kingma and Jimmy Ba},
      year={2017},
      eprint={1412.6980},
      archivePrefix={arXiv},
      primaryClass={cs.LG},
      url={https://arxiv.org/abs/1412.6980}, 
}

@misc{smith2018superconvergencefasttrainingneural,
      title={Super-Convergence: Very Fast Training of Neural Networks Using Large Learning Rates}, 
      author={Leslie N. Smith and Nicholay Topin},
      year={2018},
      eprint={1708.07120},
      archivePrefix={arXiv},
      primaryClass={cs.LG},
      url={https://arxiv.org/abs/1708.07120}, 
}

@article{agostinelli2019solving,
  title={Solving the Rubik’s cube with deep reinforcement learning and search},
  author={Agostinelli, Forest and McAleer, Stephen and Shmakov, Alexander and Baldi, Pierre},
  journal={Nature Machine Intelligence},
  volume={1},
  number={8},
  pages={356--363},
  year={2019},
  publisher={Nature Publishing Group UK London}
}

@article{chen2025iastar,
  title = {{iA*}: Imperative Learning-based {A*} Search for Path Planning},
  author = {Chen, Xiangyu and Yang, Fan and Wang, Chen},
  journal = {IEEE Robotics and Automation Letters (RA-L)},
  year = {2025},
  volume = {10},
  number = {12},
  pages = {12987-12994},
  url = {https://arxiv.org/abs/2403.15870},
  code = {https://github.com/sair-lab/iAstar},
  website = {https://sairlab.org/iastar/},
  addendum = {Highlighted as an iSeries Article in Path Planning}
}

@inproceedings{ijcai2024p743,
  title     = {Guiding GBFS through Learned Pairwise Rankings},
  author    = {Hao, Mingyu and Trevizan, Felipe and Thiébaux, Sylvie and Ferber, Patrick and Hoffmann, Jörg},
  booktitle = {Proceedings of the Thirty-Third International Joint Conference on
               Artificial Intelligence, {IJCAI-24}},
  publisher = {International Joint Conferences on Artificial Intelligence Organization},
  editor    = {Kate Larson},
  pages     = {6724--6732},
  year      = {2024},
  month     = {8},
  note      = {Main Track},
  doi       = {10.24963/ijcai.2024/743},
  url       = {https://doi.org/10.24963/ijcai.2024/743},
}

@inproceedings{orseau2023ltscm,
  title     = {Levin Tree Search with Context Models},
  author    = {Orseau, Laurent and Hutter, Marcus and Lelis, Levi H. S.},
  booktitle = {Proceedings of the Thirty-Second International Joint Conference on
               Artificial Intelligence, {IJCAI-23}},
  publisher = {International Joint Conferences on Artificial Intelligence Organization},
  editor    = {Edith Elkind},
  pages     = {5622--5630},
  year      = {2023},
  month     = {8},
  note      = {Main Track},
  doi       = {10.24963/ijcai.2023/624},
  url       = {https://doi.org/10.24963/ijcai.2023/624},
}

@inproceedings{NEURIPS2024_bc8b2058,
 author = {Zhao, Dengwei and Tu, Shikui and Xu, Lei},
 booktitle = {Advances in Neural Information Processing Systems},
 doi = {10.52202/079017-3309},
 editor = {A. Globerson and L. Mackey and D. Belgrave and A. Fan and U. Paquet and J. Tomczak and C. Zhang},
 pages = {104138--104179},
 publisher = {Curran Associates, Inc.},
 title = {{SeeA*} : Efficient Exploration-Enhanced {A*}  Search by Selective Sampling},
 url = {https://proceedings.neurips.cc/paper_files/paper/2024/file/bc8b2058fd96978a4146f18298cb2d39-Paper-Conference.pdf},
 volume = {37},
 year = {2024}
}
